\documentclass[preprint,12pt,numbers,sort&compress]{elsarticle}
\usepackage[utf8]{inputenc}
\usepackage[T1]{fontenc}
\usepackage{amsmath,amssymb,amsfonts}   
\usepackage{mathtools}                  
\usepackage{bm}                         
\usepackage{graphicx}                   
\graphicspath{{figures/}{./}}           
\usepackage{booktabs}                   
\usepackage{multirow}                   
\usepackage{array}                      
\usepackage{tabularx}                   
\usepackage{longtable}                  
\usepackage[labelfont=bf,font=small,labelsep=period]{caption}  
\usepackage{subcaption}                 
\usepackage[flushmargin]{footmisc}

\newenvironment{promptquote}{%
  \begin{quote}\small\sloppy\emergencystretch=3em%
  \setlength{\parskip}{3pt}\setlength{\parindent}{0pt}%
}{%
  \end{quote}%
}

\usepackage{makecell}                   
\usepackage{threeparttable}             

\usepackage{siunitx}                    
\usepackage{xcolor}
\usepackage[hidelinks]{hyperref}
\usepackage{url}

\usepackage[a4paper,margin=1in,top=0.9in,bottom=0.9in]{geometry}
\newcommand{\card}[1]{\lvert #1 \rvert}        
\newcommand{\fine}{\mathrm{fine}}

\newcommand{\feat}{\mathrm{feat}}

\journal{Automation in Construction}

\begin{document}

\begin{frontmatter}

\title{VLM-based automatic multi-granularity graph representation of building layouts for design informatics}

\author[mit]{Song Guo}

\author[hkust]{Zhuoshi Chen}

\author[hkust]{Maosu Li\corref{cor1}}
\cortext[cor1]{Corresponding author.}
\ead{maosuli@hkust-gz.edu.cn}

\author[tsinghua]{Weimin Zhuang}

\affiliation[mit]{%
  organization={Senseable City Lab, Department of Urban Studies and Planning, Massachusetts Institute of Technology},
  city={Cambridge},
  state={Massachusetts},
  country={United States}}

\affiliation[hkust]{%
  organization={Society Hub, The Hong Kong University of Science and Technology (Guangzhou)},
  city={Guangzhou},
  country={China}}
  
\affiliation[tsinghua]{%
  organization={School of Architecture, Tsinghua University},
  city={Beijing},
  country={China}}
\begin{abstract}
Architectural floorplan images encode rich relational knowledge among functional spaces, which underpins design retrieval, knowledge-based reasoning, and BIM enrichment through the building lifecycle. However, it remains challenging to automatically construct task-adaptive graph representations for public buildings. To address this gap, we first define a multi-granularity Level-of-Graphs (LoGs) for public building layouts. Methodologically, we present a Vision-Language Model (VLM)-based automatic LoG construction through node identification, edge inference, text parsing, and graph coarsening. VLM-generated representations are systematically evaluated and tested in real-world tasks, using 147 academic library floorplans worldwide as a case study. Experiments showed VLM-generated graphs were broadly consistent with human-labeled graphs (matched node ratio >= 92\%; 509.3 s per floor plan for three-LoG graph generation). Meso-grained graphs yield the best node-level zone prediction (Macro F1 = 0.647, at 65\% of fine-grained complexity), while coarse-grained graphs are most effective for graph-level layout quality evaluation (Spearman's~$\rho$ = 0.610, at 16\% of fine-grained complexity). By enabling scalable, annotation-free extraction of structured layout information from floorplan images, this study advances design informatics by converting plan images into knowledge representations, thereby enhancing the utilization of design information across the building life cycle.
\end{abstract}


\end{frontmatter}


\section{Introduction}
\label{sec:intro}

Design informatics enhances the efficiency and rationality of design by converting materials, such as architectural drawings, into machine-readable information, thereby facilitating the flow of design knowledge. Its core challenge lies in extracting and formalizing information from heterogeneous sources to support knowledge-intensive tasks. In the context of architectural design, building layout serves as the structural backbone of space, defining its fundamental configuration, while elements such as façades are developed based on and constrained by this layout. Building layouts encode structured relations among functional areas \citep{hillier1999space,hillier1989social}, which constitute a fundamental form of engineering knowledge in the AEC domain. Different from natural images, which consist of continuous pixels expressing texture and visual details, building layouts are essentially discrete \citep{luo2022floorplangan,ZHAO2025103396} encoding relational information \citep{SUTER2014395}, such as adjacency and connectivity. 

Graph, as a relational data structure \citep{xu2017representing}, is capable of representing building layouts' structure \citep{sangawong2019modified}, and serves as a natural carrier for downstream engineering informatics tasks such as design retrieval, knowledge-based reasoning, and BIM enrichment. Graph-based layout representations, such as bubble diagrams \citep{laseau2000graphic,pena1969problem}, have long been regarded as fundamental tools for design knowledge representation and analysis. Despite their importance, layout graphs remain challenging to automatically extract from raw floorplan images, particularly those of public buildings. Compared with residential layouts, public buildings typically contain more complex functional semantics \citep{cheon2013spatial,keles2025accessibility}, a large number of spatial zones, and more diverse partitioning forms, all of which complicate automatic recognition and representation. These challenges create a persistent information bottleneck between unstructured visual design artifacts and the structured knowledge bases required by engineering informatics applications.

The complexity of public buildings further implies that a single graph representation is not a one-size-fits-all solution for all architectural tasks. For example, the choice of representation has been shown to fundamentally determine downstream task performance in machine learning \citep{goodfellow2016deep,RICHTER2010180}, while graph-level representation expressiveness directly constrains what GNN-based models can learn \citep{xu2019powerful,WETTEWA2024102868}. Multi-level task-appropriate representations are therefore essential for building knowledge bases that serve heterogeneous downstream applications. Multi-granularity graph representations have been explored in various domains, such as image processing \citep{yang2024adaptive}, citation networks \citep{zhao2025grain}, and network attack analysis \citep{liu2025mgf}.  Particularly, in urban informatics, CityGML \citep{groger2012citygml} defines a standardized Level of Detail (LoD) concept from LoD 0 (footprint) to LoD 4 (interior), enabling different exterior component modeling of 3D city models. However, to our best knowledge, there exists no multi-granularity modeling of interior building layouts to understand the multi-level abstraction of the building layout knowledge. 

Recent advances in vision--language models (VLMs) \citep{gemini2025gemini,openai2024gpt4} provide a new direction for addressing this information bottleneck. VLMs have demonstrated remarkable capabilities in visual inference and spatial reasoning \citep{liu2023visual,stogiannidis2025mind}. In zero-shot settings, VLMs can infer relative spatial locations \citep{cai2025spatialbot}, enabling the extraction of relational information among entities in visual scenes \citep{chen2024spatialvlm}. These capabilities align directly with the key components of automatic layout graph extraction: recognizing functional elements and inferring spatial relationships between them without requiring task-specific training data. VLMs therefore provide a promising foundation for automatically constructing structured architectural knowledge representations at scale.

In this study, we propose an automatic multi-granularity graph representation for complex public building layouts, using academic libraries as a case study. We introduce fine-, meso-, and coarse-grained graph representations that can represent public building layouts at different levels of detail. To automatically and adaptively construct the representations, we propose a four-step VLM-based pipeline including node identification, edge inference, text parsing, and bottom-up coarsening, enabling fully automatic and zero-annotation extraction of complex layouts. The proposed representations are evaluated along two dimensions: fidelity with respect to human-labeled graphs, and effectiveness on representative downstream analysis tasks.

This study makes three contributions.
\begin{itemize}
\item First, it introduces a Level-of-Graphs (LoGs) for building layouts, with metrics that can be used to evaluate the fidelity and effectiveness of the graph representations, establishing a principled multi-granularity representation foundation.
\item Second, it proposes an automatic VLM-based pipeline to construct the multi-granularity representation directly from layout images, eliminating the need for manual annotation and providing a scalable route from unstructured visual artifacts to structured engineering knowledge.
\item Third, it provides empirical evidence that optimal graph granularity differs across architectural analysis tasks, contributing new knowledge on granularity--task alignment and advancing design informatics.
\end{itemize}

The remainder of this paper is organized as follows. Section~\ref{sec:related} reviews related work on graph granularity in architecture, automatic layout representation, and VLMs in architectural automation. Section~\ref{sec:methods} describes the proposed methodology of multi-granularity graph representation. Section~\ref{sec:experiments} presents the experiment settings for evaluating the multi-granularity graph representations. Section~\ref{sec:results} presents experimental results. Section~\ref{sec:discussion} discusses the significance, key findings and the limitations and future directions. Section~\ref{sec:conclusion} concludes the paper.

\section{Related Works}
\label{sec:related}

This section reviews graph granularity in architectural analysis, automatic representation of building layouts, and the application of VLMs in architectural automation.

\subsection{Graph granularity in architectural analysis}
\label{sec:related-granularity}

\begin{figure}[!htbp]
\centering
\includegraphics[width=\linewidth]{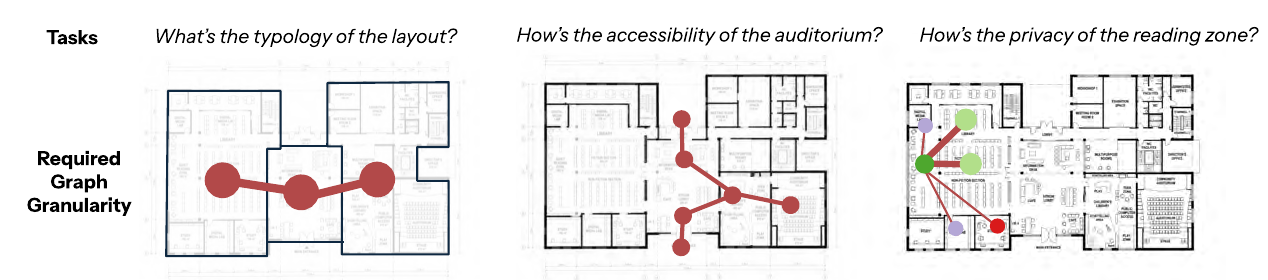}
\caption{Architectural analysis tasks require layout graphs at different granularities.}
\label{fig:granularity}
\end{figure}

Architectural analysis tasks require layout representations at different granularities, as illustrated in Figure~\ref{fig:granularity}. For whole-layout analyses such as recognizing typologies, a coarse-grained representation that captures major spatial divisions is sufficient and computationally efficient. For analyses such as evaluating accessibility of functional areas, a meso-grained representation preserving connectivity is required. For analyses related to detailed perceptions, only a fine-grained representation that resolves furniture arrangement can support meaningful inference. No single granularity dominates across tasks.

The concept of multi-granularity representation is already established in adjacent fields. In urban informatics, the CityGML standard defines five Levels of Detail (LoD 0--4) that progressively refine three-dimensional building modelling from footprint to interior \citep{groger2012citygml, biljecki2016applications}. In computer graphics, multi-scale mesh and shape representations enable progressive rendering and analysis at varying resolutions \citep{garland1997surface}. These multi-granularity frameworks share a common principle: a single physical entity can be represented at multiple abstraction levels, and the appropriate level depends on the downstream tasks. 

However, there exists no definition of the level of details in layout representation, lacking multi-level specifications of interior configurations \citep{BILJECKI201625}. Building layout graphs have largely remained at a single granularity. The choice of granularity is typically determined by the dataset annotation protocol rather than by the requirements of the downstream task, which limits the reusability of a given graph across different purposes. Empirical evidence on which granularity best supports which task is largely absent from the previous architectural studies, motivating the systematic granularity-task alignment study presented in this paper.

\subsection{Automatic representation of building layout}
\label{sec:related-layout}

\begin{figure}[!htbp]
\centering
\includegraphics[width=\linewidth]{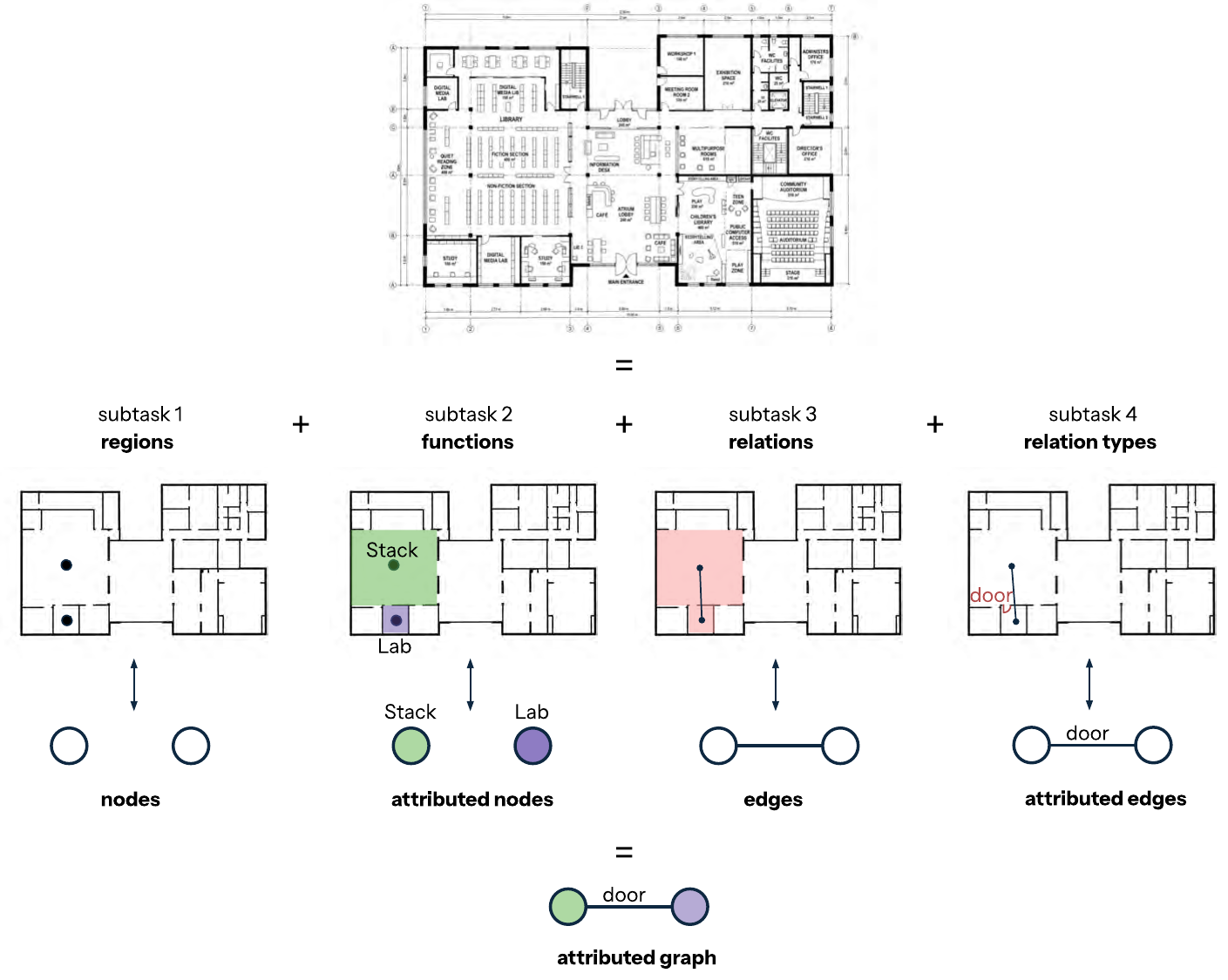}
\caption{Subtasks of automatic building layout representation in conventional approaches.}
\label{fig:subtask}
\end{figure}

Automatic representation of floor plans is challenging because architectural drawings contain heterogeneous information across multiple modalities \citep{xing2026multimodal,WANG2025103278} and scales \citep{xu2024multiscale}. Unlike Building Information Modeling (BIM), which contains explicitly structured component information \citep{LANGENHAN2013413}, a typical floor plan image includes geometric shapes, semantic elements, and textual annotations \citep{xu2025automatic,xu2024multiscale}. Under a deep learning paradigm, layout interpretation is usually modular \citep{SCHONFELDER2025103761} and can decompose into at least four interdependent subtasks: identifying regions, their functions, their relations, an`d relation types (Figure ~\ref{fig:subtask}). Each subtask further depends on coordinated detection and interpretation of multiple object types \citep{shteriyanov2025blueprintsymvl,yang2019semantic}. For example, relation identification requires the detection of walls, doors, and windows, each of which may be a separately trained deep learning model.

\begin{table}[!htbp]
\centering
\caption{Comparison of layout graph construction approaches for architectural floor plans.}
\label{tab:method-compare}
\scriptsize
\setlength{\tabcolsep}{2.5pt}
\renewcommand{\arraystretch}{1.2}
\begin{tabularx}{\textwidth}{@{}>{\raggedright\arraybackslash}p{1.3cm} >{\raggedright\arraybackslash}p{1.2cm} >{\raggedright\arraybackslash}X >{\raggedright\arraybackslash}X >{\raggedright\arraybackslash}X >{\raggedright\arraybackslash}X >{\raggedright\arraybackslash}p{1.0cm} >{\raggedright\arraybackslash}p{1.1cm} >{\raggedright\arraybackslash}p{1.0cm} >{\raggedright\arraybackslash}p{1.0cm}@{}}
\toprule
\textbf{Approach} & \textbf{Ref.} & \textbf{Region} & \textbf{Function} & \textbf{Relation} & \textbf{Relation type} & \textbf{Annot.\ dep.} & \textbf{Integ.} & \textbf{Granul.} & \textbf{Public-bldg} \\
\midrule
\multirow{6}{*}{Manual}
   & \citep{lu2025complex}              & Manual & Manual & Positional adjacency & Learned embedding & High & Separate & Single & Yes \\
   & \citep{xin2025prompts}             & Manual & Manual & Manual & Not applicable & High & Separate & Single & Partial \\
   & \citep{HUANG2025111922}            & Manual & Not applicable & Manual & Not applicable & High & Separate & Single & Yes \\
   & \citep{lin2024edge,weber2024hypergraph} & Manual & Manual & Manual & Defined by connecting component & High & Separate & Single & No \\
   & \citep{guo2026function}            & Manual & Manual & Labeled connecting component & Defined by connecting component & High & Separate & Single & Yes \\
   & \citep{wang2023automated}          & Manual & Rule-based & Positional adjacency & Not applicable & Medium & Separate & Single & No \\
\midrule
\multirow{4}{*}{\shortstack[l]{Deep\\learning}}
   & \citep{azizi2022graph,maeng_data-driven_2021} & Image segmentation & Manual & Defined by opening or adjacency & Not applicable & Medium & Separate & Single & No \\
   & \citep{knechtel2024semantic}       & Image segmentation & Supervised learning & Positional adjacency + door detection & Defined by adjacency or door & Low & Separate & Single & No \\
   & \citep{yang_representation_2023}   & Binarized image & Text OCR & Positional adjacency & Defined by node function pair & Low & Separate & Single & Yes \\
   & \citep{dong2021vectorization}      & GAN-based vectorization & GNN inference & Labeled connecting component & Defined by connecting component & Medium & Separate & Single & No \\
\midrule
\multirow{1}{*}{\textbf{\shortstack[l]{VLM\\zero-shot}}}
   & \textbf{Ours} & \textbf{VLM zero-shot} & \textbf{VLM zero-shot} & \textbf{VLM zero-shot} & \textbf{VLM zero-shot} & \textbf{None} & \textbf{Full} & \textbf{Multi} & \textbf{Yes} \\
\bottomrule
\end{tabularx}
\end{table}
Existing building layout graph construction approaches are primarily based on manual annotation or partially automatic deep learning methods, whereas fully automatic approaches enabled by VLMs remain scarce (Table \ref{tab:method-compare}). Manual approaches rely on human annotators to trace region contours \citep{weber2024hypergraph}, assign functional labels \citep{guo2026function}, and mark either the connecting components between rooms \citep{dong2021vectorization} or the relationships \citep{xin2025prompts}. Partially automatic approaches extract regions from floor plans through deep learning–based segmentation \citep{azizi2022graph,maeng_data-driven_2021,knechtel2024semantic} or other vectorization methods \citep{yang_representation_2023,dong2021vectorization}, and assign functional labels to regions via supervised training \citep{knechtel2024semantic}, or model inference \citep{dong2021vectorization}. Connections are established from positional adjacency between regions \citep{azizi2022graph,maeng_data-driven_2021}, or further given semantic types based on the detection of connecting components such as doors \citep{knechtel2024semantic,dong2021vectorization}. Partially automatic methods typically work only for residential buildings, where training data is sufficient and design rules are well explicit, and their performance on public buildings remains lower \citep{yang_representation_2023}. VLM-based approaches offer the opportunity to recognize floor plans within a single call \citep{defazio2025vision}. Nevertheless, apart from the framework proposed in this paper, no existing VLM-based method performs complete identification of regions, functions, and relations with fine-grained semantics in a unified pipeline. 

Across all surveyed approaches, four main gaps remain. First, they rely heavily on annotated datasets, yet existing publicly available floor-plan datasets have only limited coverage of public buildings \citep{lu2025complex} and largely depend on manual annotation \citep{guo2026function,xu2025automatic,khan2025generative}. Second, they lack pipeline integration thus hindering the computational convenience, requiring separate models or manual interventions for the subtasks, which fragments the workflow and limits end-to-end automation. Third, they lack flexibility without an adaptive representation framework that can adjust granularity to different downstream tasks. Fourth, they generalize poorly to complex public buildings when explicit visual cues are incomplete, especially for open-plan areas or curved boundaries \citep{knechtel2024semantic}. 
The framework proposed in this paper addresses these gaps by integrating all four subtasks into a single VLM-based zero-shot pipeline that requires no task-specific training data, and operates directly on complex public-building floor plans.

\subsection{VLMs in architectural automation}
\label{sec:related-vlm}

Large language models (LLMs) and VLMs have been increasingly leveraged in architectural automation \citep{ERFANI2026103909}, spanning the construction phase and the design phase. In the construction phase, VLMs have been applied to vision-based monitoring \citep{hussain2026vision, sivanraj2026real}, occupation analysis \citep{qaisar2026exploring}, and safety risk detection \citep{cheng2024computer, wang2026integrating,WANG2026103985}. These applications explore the ability of pretrained vision-language backbones to recognize scene semantics. In the design phase, computational design studies have explored LLMs for iterative prompt-based workflows in parametric modelling \citep{khan2025generative}, where natural language descriptions are converted into geometric constraints. VLMs have been applied to infer room functions by reasoning jointly over detected room geometries and visual context \citep{zong2025housetune}, demonstrating that vision-language joint reasoning can substitute for hand-crafted geometric features in building layouts. These early applications suggest that the multimodal reasoning capability of VLMs has the potential to align with the heterogeneous nature of architectural drawings.

For layout generation, generative models combined with LLMs and VLMs reduce reliance on large annotated datasets \citep{xin2025prompts}, though most existing approaches still require graphical inputs such as boundary shapes \citep{shim2024floordiffusion}, spatial relationships \citep{shabani2022housediffusion}, or sketches \citep{leng2024archidiffusion} to ground the generation. For spatial understanding, VLMs have been applied to parse floor plans \citep{defazio2025vision} and to extract spatial knowledge from 3D simulated environments for robot navigation \citep{sun2025enhancing}, in which VLMs have demonstrated that zero-shot inference can recover room-level connectivity without training data \citep{defazio2025vision}. However, these applications are primarily focused on limited building typologies. Performance notably degrades in large open areas \citep{defazio2025vision}. This may be due to: (i) the inherent complexity of public building layouts; (ii) the limited availability of validation datasets; and (iii) the variability and uncertainty in VLM performance under different query and answer strategies.

Overall, existing automatic layout representation approaches remain limited by their focus on residential buildings and dependence on manual annotation, while VLMs' potential for fine-grained spatial relationship extraction remains underexplored. To address these gaps, we propose an automatic VLM-based pipeline for multi-granularity graph extraction and introduce the Level-of-Graphs (LoGs), empirically establishing which granularity level best supports which architectural analysis task.

\section{Multi-granularity Graph Representation}
\label{sec:methods}
This section focuses on the methodological contributions independent of the particular dataset and downstream tasks. We first introduce the Level-of-Graphs (LoGs), then describe the automatic VLM-based pipeline, and finally present a set of general evaluation metrics that characterize the constructed graph's fidelity, effectiveness and computational complexity on downstream tasks. 

\subsection{The LoGs}

\label{sec:methods-log}
We propose a three-level LoG representation that can describe an architectural floor plan at different granularities. Whereas CityGML LoD~0--4 \citep{groger2012citygml} describes 3D buildings from a modeling perspective, the LoGs formalize the graph representation of interior spatial configuration (Figure~\ref{fig:lod}). The three levels are defined as follows:

\begin{figure}[!htbp]
\centering
\includegraphics[width=\linewidth]{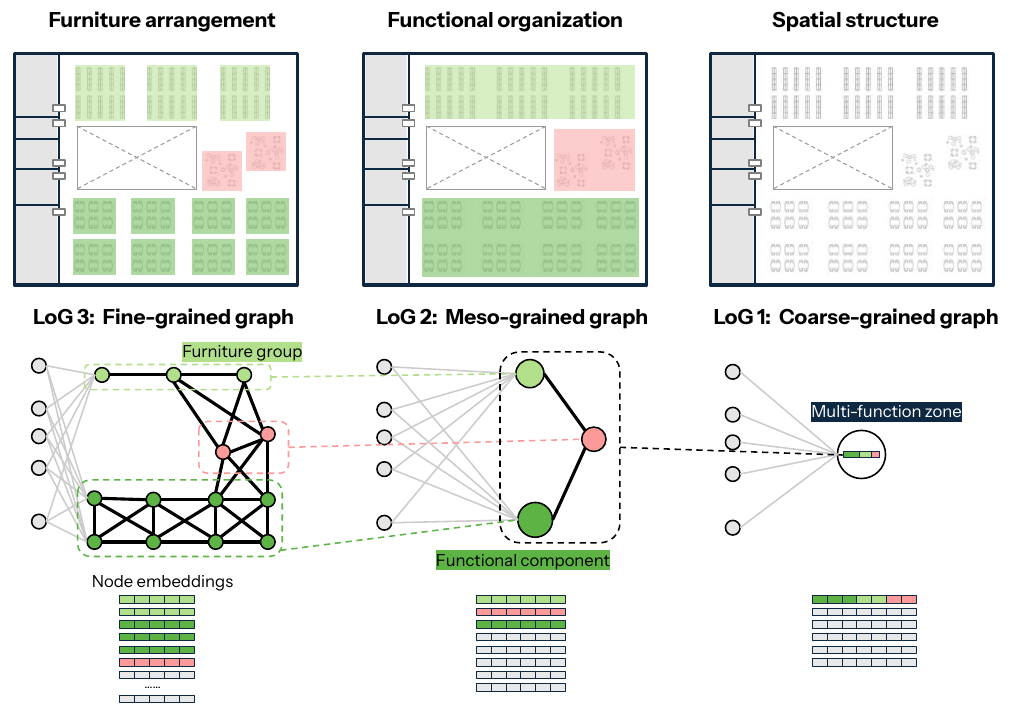}
\caption{The LoGs for building layout representation.}
\label{fig:lod}
\end{figure}
\begin{itemize}
\item \textbf{LoG~1 (coarse-grained):} each node represents a maximal contiguous spatial region enclosed by physical boundaries (i.e., a room). A contiguous region may contain multiple functional components. It provides an objective representation of the building’s physical spatial partitioning, independent of functional allocation and furniture arrangement. This level is typically fixed after construction and only changes under conditions such as renovation, extension, or flexible subdivision and merging of rooms.
\item \textbf{LoG~2 (meso-grained):} each node represents a contiguous functional component. A functional component may contain multiple similar furniture groupings. It reflects the functional distribution within the layout. This level may vary with changes in spatial use, but remains independent of the size and spatial distance of individual furniture groupings.
\item \textbf{LoG~3 (fine-grained):} each node represents a furniture grouping (e.g., a group of reading seats, a stack row). It captures spatial units formed by furniture arrangement, exhibiting scale-sensitive characteristics. This level is the most dynamic and may change with variations in furniture layout.

\end{itemize}

In terms of application scenarios, LoG~1 is primarily used to capture the overall properties and physical allocation of building layouts, such as characterizing or predicting footprint shape, massing, and basic typologies, as well as allocating the area ratios of different room types. LoG~2 is suited for analyzing functional aspects, including functional programming and distribution, as well as the proximity and accessibility among functional spaces. LoG~3 is used to investigate the relationship between layout and user behavior and perception, such as occupancy capacity, the alignment between layouts and user requirements, and the spatial influence on users’ physical and psychological experiences. Within AEC workflows, LoG~1 is mainly used by construction and structural engineers, LoG~2 by architects, and LoG~3 by interior designers and facility managers.

\subsection{VLM-based automatic graph construction}
\label{sec:methods-construct}
\begin{figure}[!htbp]
\centering
\includegraphics[width=\linewidth]{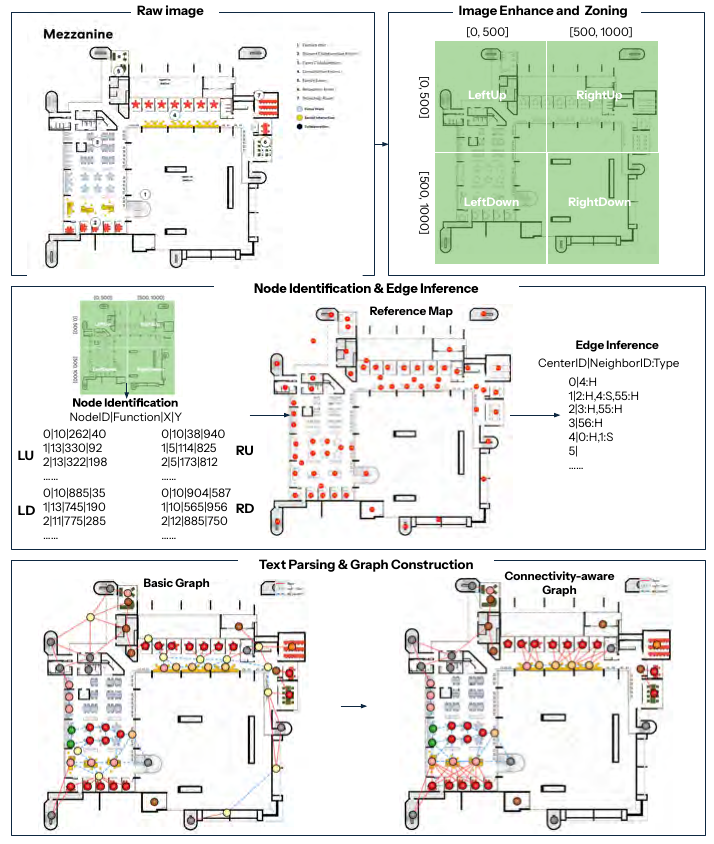}
\caption{automatic graph construction pipeline.}
\label{fig:pipeline}
\end{figure}
To fulfill the automatic construction of the multi-granularity graphs, we design a VLM-based pipeline. Graph construction is decomposed into four stages: node identification, edge inference, text parsing, and bottom-up coarsening (Figure~\ref{fig:pipeline}). The first three stages produce the LoG~3 (fine-grained) graph directly from floor plan images. The fourth stage derives the LoG~2 (meso-grained) and LoG~1 graphs (coarse-grained) via the coarsening procedure.

\subsubsection{Node identification}
\label{sec:methods-node}
The floor plan images are first enhanced using Gemini \citep{gemini2025gemini} to improve their readability for VLM recognition. Distracting annotations such as numbers and text labels are removed, while partially occluded elements are inferred and reconstructed. Image resolution and contrast are also improved. Node identification is then performed in four spatial passes over the same enhanced floor plan: in each pass, the VLM analyzes the complete image but is instructed to output only nodes whose centers fall within one of four spatial regions (LeftUp, RightUp, LeftDown, RightDown quadrants). This region-filtered strategy improves detection completeness and positional accuracy compared to single-pass processing. 

Within each pass, the VLM is required to identify all fine-grained functional spaces (see ~\ref{app:node-prompt}). A node is defined as a furniture group rather than a walled room: boundaries are inferred from furniture symbols, functional zoning, textual labels, partitions, and global layout. Each identified node is assigned a functional category from a predefined set\footnote{Classes: 0---Courtyard/Terrace, 1---Information/Service, 2---Collections, 3---Reading Seats/Learning Commons, 4---Lounge/Recreation, 5---Group Study, 6---Self-study/Individual Study, 7---Classroom, 8---Auditorium, 9---Specialized Room/Studio/Workshop/Exhibition, 10---Stair/Elevator/Escalator, 11---Restroom/Toilet, 12---Corridor, 13---Meeting/Office/Others. Functional descriptions and plan examples can be found in \citep{guo2026function}.} defined by prior investigations \citep{guo2026function,guo2024enhancing}. Identified nodes are merged across regions by deduplicating spatially proximate detections, producing a reference map annotated with node IDs and positions.

\subsubsection{Edge inference}
\label{sec:methods-edge}
After node identification, edge inference is performed in a node-centered iterative manner rather than reasoning over all nodes simultaneously. In each iteration, the VLM receives two images, the reference map with detected nodes marked as dots with IDs, and the original enhanced floor plan providing complete visual context of partitions and connecting components. The model is asked to identify the directly adjacent neighbors of one designated center node (see ~\ref{app:edge-prompt}). Two spaces are deemed directly adjacent only if they are immediate local neighbors and the transition between them does not itself constitute a distinct functional space. Long-distance edges across halls, edges through intermediate zones, and edges based merely on co-membership in an open-plan room are explicitly prohibited. Each adjacency is further typed as a hard connection (``H'') through a door, or a soft connection (``S'') without any partitions.

\subsubsection{Text parsing and graph construction}
\label{sec:methods-parse}
VLM outputs are produced as plain text to reduce computational cost. They are parsed to extract node attributes (ID, functional category, normalized coordinates) and edges (center ID, neighbor ID,edge type), with duplicate edges across iterations removed. Two types of graphs are then constructed following the definitions in \citep{guo2026function}: the Basic Graph (BG) preserves all nodes and inferred connections, including corridor nodes, thereby reflecting the original layout, while the Connectivity-aware Graph (CaG) removes corridor nodes and reconstructs connectivity across functional areas through graph transformation. All subsequent analyses adopt the CaG representation as the LoG~3 (fine-grained) graph.

\subsubsection{Graph coarsening}

The LoG~2 (meso-grained) and LoG~1 (coarse-grained) graphs are then derived from the LoG 3 (fine-grained) graph through bottom-up coarsening. For LoG~2, LoG~3 nodes sharing the same functional category and connected via adjacent edges are merged into a single LoG~2 node. For LoG~1, all LoG~3 nodes connected via adjacent edges are merged regardless of functional category. To preserve semantic information that would otherwise be lost during coarsening, each LoG~1 node stores a function composition dictionary recording the member count of each functional category among its constituent LoG 3 nodes, with the dominant function assigned by majority vote. This is designed to be suited for open-plan spaces containing multiple functions. Edges in LoG~1 and LoG~2 graphs are aggregated by majority vote over their constituent LoG~3 edges, with ties broken in the order of adjacent, door, corridor. 

\subsection{Graph evaluation metrics}
\label{sec:methods-metrics}
 To support systematic evaluation of the constructed multi-granularity graphs, we apply graph fidelity metrics and computational complexity measures. 

\subsubsection{Graph fidelity metrics}
\label{sec:methods-fidelity}

\begin{figure}[!htbp]
\centering
\includegraphics[width=\linewidth]{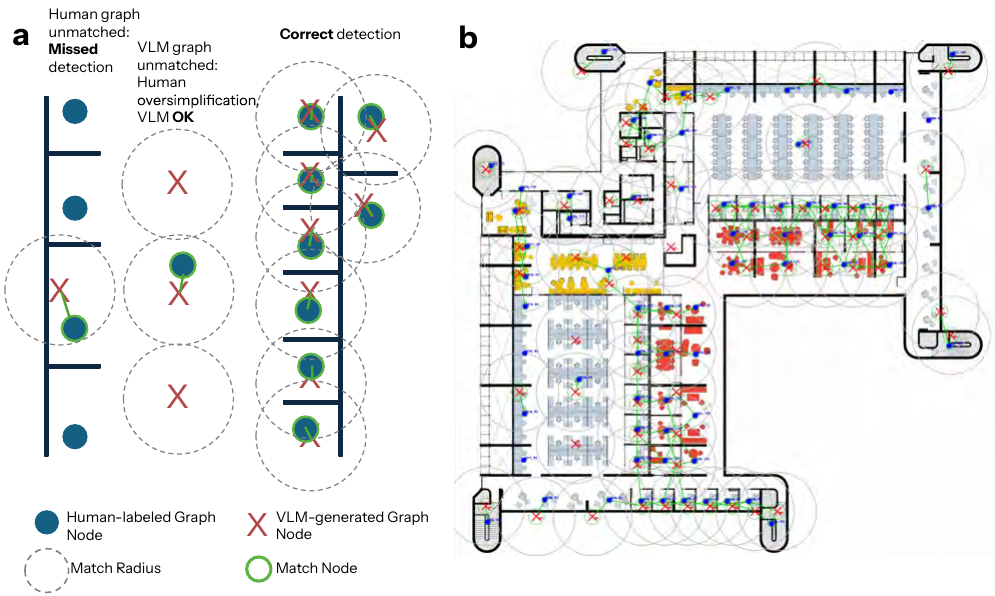}
\caption{Node matching procedure between VLM-generated and human-labeled graphs. (a) Matching principle. (b) One matching example.}
\label{fig:matching}
\end{figure}

Fidelity is characterized by the structural similarity between a constructed graph and a reference graph (e.g., a human-labeled graph). Because the nodes in VLM-generated and reference graphs are not in one-to-one correspondence, node and edge matching must be established before similarity can be computed. At the LoG~3 (fine-grained) level, nodes are matched by coordinate-space proximity (Figure~\ref{fig:matching}): for each VLM node, all reference-graph nodes whose aligned coordinates fall within a predefined radius are recorded as candidate matches, reflecting the extent to which the reference graph is covered by the VLM-generated graph. At coarser levels, matching is inherited through membership: a super-node in one graph is matched to the super-node in the other graph with the greatest overlap in constituent fine-level members. Edges are considered matched when both of their endpoints are matched.

Given the established matching, similarity can be evaluated across four aspects: basic properties (node and edge count distributions), alignment (the fractions of matched nodes and edges), semantic consistency (functional category agreement and edge type agreement between matched pairs), and topological consistency (structural measures such as graph density and average path length between corresponding graphs). Table~\ref{tab:similarity} summarizes the metrics.
 
{\footnotesize
\setlength{\tabcolsep}{4pt}
\renewcommand{\arraystretch}{1.4}
\begin{longtable}{@{}>{\raggedright\arraybackslash}p{1.8cm} >{\raggedright\arraybackslash}p{3.0cm} >{\raggedright\arraybackslash}p{5.5cm} >{\raggedright\arraybackslash}p{4.5cm}@{}}
\caption{Similarity metrics for multi-granularity graph fidelity comparison.}\label{tab:similarity}\\
\toprule
\textbf{Aspect} & \textbf{Metric} & \textbf{Formula} & \textbf{Notes} \\
\midrule
\endfirsthead
\toprule
\textbf{Aspect} & \textbf{Metric} & \textbf{Formula} & \textbf{Notes} \\
\midrule
\endhead
\bottomrule
\multicolumn{4}{r}{\textit{(continued on next page)}}\\
\endfoot
\bottomrule
\endlastfoot
 
Basic properties
& Node count similarity
& $S_n = 1 - \dfrac{\lvert \card{V_H} - \card{V_V} \rvert}{\max(\card{V_H}, \card{V_V})}$
& $\card{V_H}$, $\card{V_V}$: node counts of the reference (human-labeled) and constructed (VLM-generated) graphs \\[6pt]
 
& Edge count similarity
& $S_e = 1 - \dfrac{\lvert \card{E_H} - \card{E_V} \rvert}{\max(\card{E_H}, \card{E_V})}$
& $\card{E_H}$, $\card{E_V}$: edge counts \\[6pt]

\midrule
Alignment
& Matched node ratio
& $R_n = \dfrac{\card{M_H}}{\card{V_H}}$
& $\card{M_H}$: number of reference-graph nodes with at least one VLM match; measures how much of the reference graph is covered \\[6pt]
 
& Matched edge ratio
& $R_e = \dfrac{\card{M^E_H}}{\card{E_H}}$
& $\card{M^E_H}$: number of reference-graph edges whose both endpoints are matched to VLM nodes \\[6pt]
\midrule
 
Semantic consistency
& Node function distribution similarity
& $S_f = \dfrac{\bm{f}_H \cdot \bm{f}_V}{\lVert \bm{f}_H \rVert \, \lVert \bm{f}_V \rVert}$
& $\bm{f}$: count vector over 13 functional categories; cosine similarity \\[6pt]
 
& Edge function pair similarity
& $S_{ep} = \dfrac{\bm{p}_H \cdot \bm{p}_V}{\lVert \bm{p}_H \rVert \, \lVert \bm{p}_V \rVert}$
& $\bm{p}$: count vector over functional category pairs connected by edges; cosine similarity \\[6pt]
\midrule
 
Topological consistency
& Graph density similarity
& $S_d = 1 - \lvert d_H - d_V \rvert$
& $d = 2\card{E} \,/\, (\card{V}(\card{V}-1))$ \\[6pt]
 
& Average path length similarity
& $S_\ell = 1 - \dfrac{\lvert \bar{\ell}_H - \bar{\ell}_V \rvert}{\max(\bar{\ell}_H, \bar{\ell}_V)}$
& $\bar{\ell}$: average shortest path length on the largest connected component \\
\end{longtable}
}

\subsubsection{Computational complexity}
\label{sec:methods-complexity}

Computational complexity is another consideration in evaluating graph representations. For both deep-learning-based and machine-learning-based tasks, the graph size directly influences processing time. We therefore apply a set of relative complexity measures independent of specific datasets or downstream tasks. These measures are reported in Section~\ref{sec:results} to support trade-off decisions between task performance and computational efficiency.

\paragraph{GNN message-passing complexity}
For GNN, the per-layer time complexity of message passing on a graph $G = (V, E)$ is \citep{xu2019powerful,hamilton2017inductive}:
\begin{equation}
\label{eq:gnn-full}
C_{\mathrm{GNN}}(G, L) \;=\; L \cdot \mathcal{O}\!\left(\card{E} \cdot d_{\mathrm{in}} \cdot d_{\mathrm{out}}\right),
\end{equation}
where $L$ is the number of message-passing layers, $\card{E}$ is the number of edges, and $d_{\mathrm{in}}$, $d_{\mathrm{out}}$ are the input and output feature dimensions per layer. When the hidden dimension is fixed across granularity and graph types, the relative computational cost of a graph at LoG $\ell$ with respect to the LoG 3 (fine-grained) baseline is therefore as follows:
\begin{equation}
\label{eq:r-gnn}
r_{\ell} \;=\; \frac{\card{E_{\ell}}}{\card{E_{LoG 3}}},
\end{equation}
where $\card{E_{\ell}}$ denotes the mean edge count at LoG $\ell$ averaged across all floor plans, and $\card{E_{LoG3}}$ is the corresponding LoG 3 (fine-grained) mean.

\paragraph{Machine learning feature extraction complexity}
For tasks that rely on pre-computed graph-level features, certain path-dependent features dominate preprocessing cost. For example, betweenness centrality and average shortest path length require $\mathcal{O}(\card{V} \cdot \card{E})$ time on unweighted graphs when computed using Brandes' algorithm \citep{brandes2001faster}. The relative feature extraction cost at LoG $\ell$ is:
\begin{equation}
\label{eq:r-feat}
r_{\ell}^{\feat} \;=\; \frac{\card{V_{\ell}} \cdot \card{E_{\ell}}}{\card{V_{LoG3}} \cdot \card{E_{LoG3}}}.
\end{equation}

\section{Experiment Settings}
\label{sec:experiments}

After proposing the generalizable approach of multi-granularity graph representation in Section~\ref{sec:methods}, we now describe our experiment settings that apply the approach to a real-world dataset and evaluate the resulting graphs along graph fidelity and task effectiveness. 

\subsection{Dataset}
\label{sec:datasets}

The proposed approach is applied to an academic library floorplan dataset collected by  \citep{guo2026function}. The dataset comprises all globally available academic library cases from the ArchDaily website \footnote{https://www.archdaily.com/}, with interior design, temporary design, and excessively small cases filtered out. Given the complexity of academic library layouts which are characterized by expansive open areas, intricate circulation spaces, and diverse, adaptive functions \citep{YANG2026103922}, this dataset provides a suitable testbed for evaluating the effectiveness of the approach. The dataset provides both the original floor plan images and human-labeled graphs corresponding to the LoG~3 (fine-grained) graph. Three floor plans with severe image degradation were excluded, resulting in 147 valid cases. 

The proposed pipeline takes only the floor plan images as input, while the human-labeled graphs serve solely as the reference for comparison. For each case, our pipeline produces three LoG graphs. The human-labeled LoG~3 graph is similarly coarsened to LoG~2 and LoG~1 to enable subsequent comparison, resulting in six graphs per case across two sources (VLM-generated, human-labeled) and three granularity levels (LoG~1, LoG~2, LoG~3).


\subsection{Fidelity evaluation: VLM-generated vs.\ human-labeled graphs at each granularity}
\label{sec:exp-fidelity}

The VLM-generated graph is normalized to a $0\text{--}1000$ space during VLM inference, while the human-labeled graph is in the original floor plan image coordinates. An alignment step is required before matching. For each case, the human-labeled graph is projected into the VLM coordinate space using control points and an affine transformation estimated via RANSAC least-squares. With both graphs in a common coordinate system, node matching at the LoG~3 (fine-grained) level is performed by coordinate-space proximity following the procedure described in Section~\ref{sec:methods-fidelity}. For each VLM node, all human-labeled nodes whose aligned coordinates fall within a radius of 300~px are recorded as candidate matches. Given the established matching at each level, the similarity metrics summarized in Table~\ref{tab:similarity} are computed for every VLM--human graph pair.

\subsection{Effectiveness evaluation: downstream task performance and complexity}
\label{sec:exp-effectiveness}

\begin{figure}[!htbp]
\centering
\includegraphics[width=\linewidth]{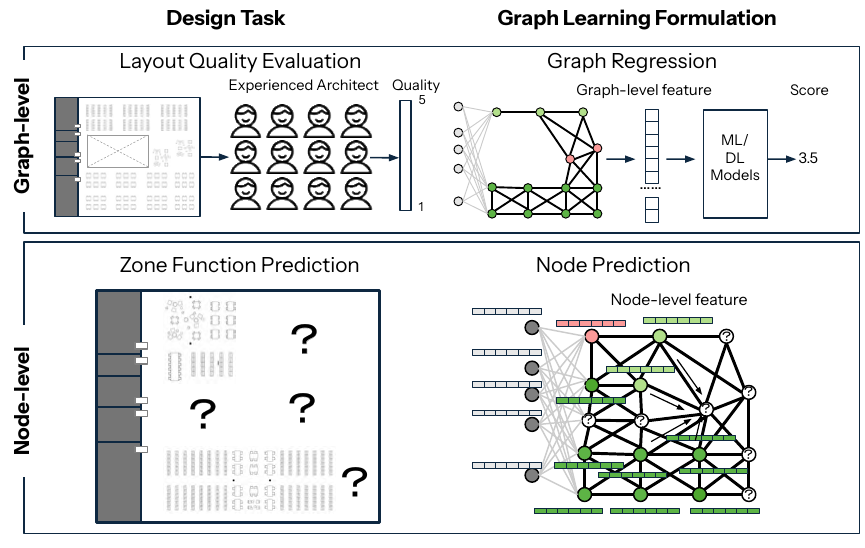}
\caption{Formulation of downstream tasks for evaluating multi-granularity graph representations.}
\label{fig:tasks}
\end{figure}

Beyond fidelity, the practical value of a graph representation lies in its ability to support downstream architectural tasks at an acceptable computational cost. Two downstream tasks are formulated (Figure~\ref{fig:tasks}). Each task is evaluated across all six combinations of graph source (VLM-generated, human-labeled) and granularity level (LoG~1--3).

To operationalize the multi-granularity graphs for downstream tasks, we apply a common feature-design rationale. Graph-level features characterize whole-layout properties (Table~\ref{tab:graph-features}), while node-level features characterize individual functional zones (Table~\ref{tab:node-features}). Seven feature groups are common to both levels: topology, degree, path, centrality, function, composition, and edge type. Node-level features additionally include spatial coordinates, since functional zones in public buildings tend to exhibit spatial regularities that are not recoverable from graph topology alone.

\subsubsection{Graph-level task: layout quality evaluation}
\label{sec:exp-regression}

During the iteration of design alternatives, evaluating layout quality is a key step for case studies and design refinement. This process can be formulated as a regression task, where expert-assessed quality scores\footnote{Quality scores were averaged across multiple experienced architects for each floor plan.} serve as the prediction target. The 42-dimensional graph-level feature vector (Table~\ref{tab:graph-features}) is extracted for each graph source--granularity combination. 

Three regression models with complementary inductive biases are tested: Support Vector Regression with RBF kernel (SVR-RBF) for nonlinear local patterns, Ridge regression for linear trends with multicollinearity control, and XGBoost for nonlinear feature interactions. Hyperparameters are selected via Leave-One-Out Cross-Validation. Evaluation uses repeated 5-fold cross-validation (10 repeats), with Spearman correlation reported as the primary performance metric. Permutation feature importance and paired permutation tests (10{,}000 permutations) are used to assess feature contributions and the significance of granularity-level differences within each graph source.

\subsubsection{Node-level task: zone function prediction}
\label{sec:exp-classification}

At the final stage of architecture design, determining the functional allocation of spatial zones is a key step. This is particularly challenging in open-plan configurations, where functional boundaries and contents are hard to define. The task is formulated as semi-supervised node classification, where the goal is to infer the functional category of target nodes given partial knowledge of surrounding node functions, mimicking real-world scenarios in which some zones are unlabeled and must be inferred from spatial context.

The prediction focuses on user-oriented functional categories within open spaces, including collections, reading areas, lounge, group study areas, and self study areas. Five functional categories (Classes~2--6) serve as prediction targets, while all other functional types are used as contextual inputs with their labels provided. Each node is represented by the 25-dimensional feature vector summarized in Table~\ref{tab:node-features}. Four representative GNN architectures are evaluated: GCN, GraphSAGE, GAT, and GINE (Graph Isomorphism Network with edge features) \citep{hu2020strategies} \footnote{Each model adopts a three-layer message-passing architecture with a hidden dimension of 64, batch normalization, and a dropout rate of 0.3. Models are trained using the Adam optimizer with an initial learning rate of $10^{-3}$ and weight decay of $10^{-4}$. A ReduceLROnPlateau scheduler is applied, and early stopping is triggered after 20 epochs without improvement on validation macro-F1. The dataset is partitioned into training, validation, and test sets at a ratio of 70:15:15 via stratified sampling on floor plan typology (courtyard, grid, or linear). All configurations were trained 5 times with identical hyperparameters and random seed (=42). Reported F1 scores are means across runs.}.Feature-level analyses are conducted on the best-performing configuration to interpret model decisions.

\section{Results}
\label{sec:results}

This section presents experimental results based on the approach in Section~\ref{sec:methods} and the experiment settings in Section~\ref{sec:experiments}. 
\begin{figure}[!htbp]
\centering
\includegraphics[width=\linewidth,height=0.85\textheight,keepaspectratio]{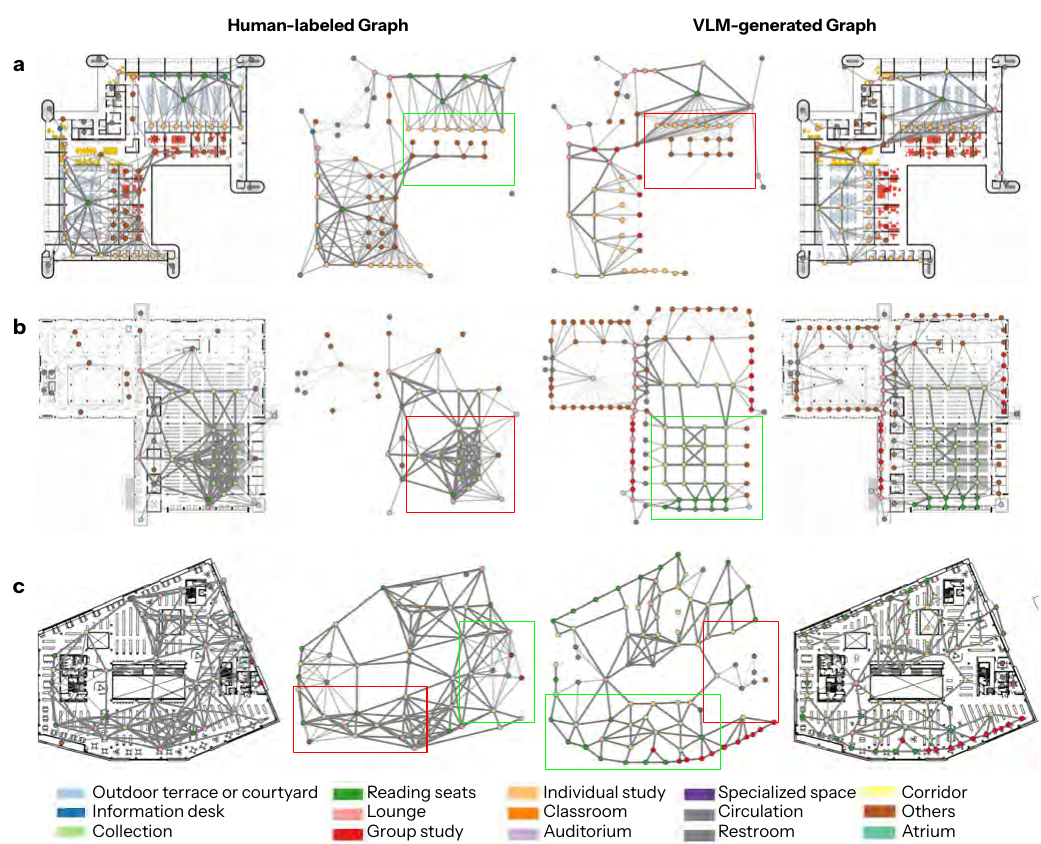}
\caption{Representative cases of human-labeled and VLM-generated graphs.}
\label{fig:cases}
\end{figure}
\subsection{Constructed multi-granularity graphs}
\label{sec:results-qualitative}

Our method is fully automated and does not require manual annotation. We use Gemini~3~Pro for VLM-based graph construction, as our experiments show that other state-of-the-art (SOTA) VLMs (Qwen~3.5, GPT~4.5, and Claude~Opus~4.6) struggle to identify and localize functional zones in complex floor plans (Figure~\ref{fig:appendix-vlm-compare}). Under batch processing, node identification required an average of 147.071 s per floor plan (approximately 1.401 s per node). Edge inference, text parsing, and LoG~3 graph construction required an average of 352.048 s per floor plan, while graph coarsening for generating LoG~2 and LoG~1 graphs required 10.204 s per floor plan. The end-to-end runtime for generating three LoGs from a raw floor plan image, without any manual annotation, averaged 509.323 s.

\begin{figure}[!htbp]
\centering
\includegraphics[width=\linewidth,height=0.85\textheight,keepaspectratio]{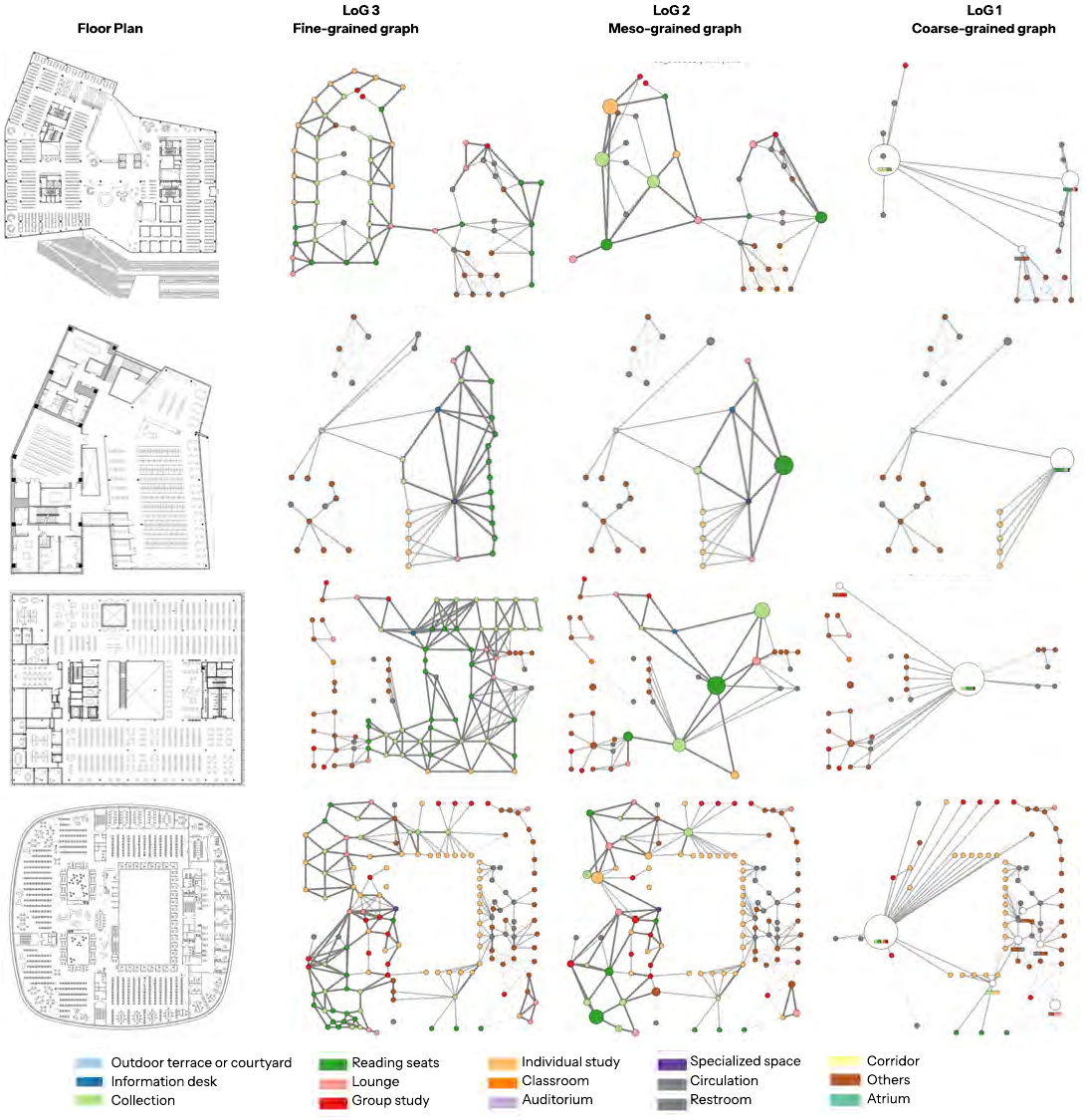}
\caption{Multi-granularity graph representations of library floor plans.}
\label{fig:lod-typology}
\end{figure}

Figure~\ref{fig:cases} illustrates examples of human-labeled and VLM-generated LoG~3 (fine-grained) graphs. Both representations capture the general spatial configuration but exhibit complementary divergences at the local details. Human-labeled graphs tend to merge spatially adjacent furniture groups sharing the same function into unified units, which reduces local spatial resolution. In regions with high concentrations of a single function, such as the dense stack and reading seat groups visible in cases~(b) and~(c), human-labeled graphs may underrepresent the internal spatial heterogeneity and limit local topological fidelity (red boxes).

VLM-generated graphs, by contrast, preserve individual furniture groups as separate nodes, yielding denser and spatially more granular representations (case~(b), green box). However, VLM-generated graphs exhibit slight deficiencies in capturing cross-zone connectivity: corridor-mediated connections spanning multiple functional regions are absent or underrepresented (red boxes, cases~(a) and~(c)). In case~(c), the VLM-generated graph correctly resolves the dense reading groups (green box) in the irregular building footprint but fails to connect them to peripheral reading areas, fragmenting what is a continuous spatial field (red box). These complementary patterns suggest that human-labeled graphs offer topologically stable representations of global spatial structure, while VLM-generated graphs provide finer local detail at the cost of reduced accuracy in recognizing multi-hop connectivity relationships.

Figure~\ref{fig:lod-typology} further illustrates examples of the multi-granularity graph representation across library layouts. The bottom-up coarsening algorithm produces well-formed graphs at each LoG level without manual tuning, demonstrating robustness to variation in building scale, geometry, and functional density. The progressive simplification from LoG~3 to LoG~1 is visually legible and architecturally interpretable, supporting the use of this multi-level representation as a foundation for downstream analysis tasks.                                  ·

\subsection{Fidelity comparison between VLM-generated and human-labeled graphs}
\label{sec:results-fidelity}

This subsection compares VLM-generated graphs with human-labeled references at all three LoG levels.

\subsubsection{LoG 3 graph comparison}
\label{sec:results-fid-fine}
\begin{figure}[!htbp]
\centering
\includegraphics[width=\linewidth,height=0.85\textheight,keepaspectratio]{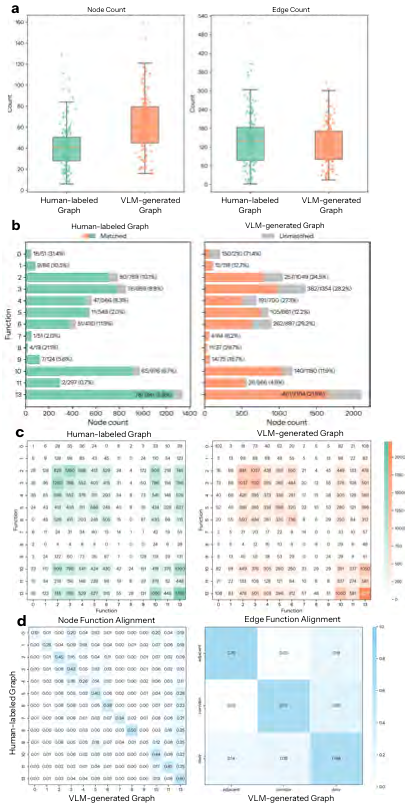}
\caption{Quantitative comparison between human-labeled and VLM-generated LoG~3 graphs across 147 library layouts. (a) Node and edge count distributions; (b) Node matching rates by functional category; (c) Edge co-occurrence matrices; (d) Node function and edge type alignment matrices over matched pairs.}
\label{fig:quant-compare}
\end{figure}
Figure~\ref{fig:quant-compare} presents a quantitative comparison at the LoG~3 (fine-grained) level. VLM-generated graphs contain more nodes than human-labeled graphs, while edge counts are comparable and slightly lower in VLM graphs. This finding is consistent with the qualitative observation that VLM representations preserve individual furniture groups as separate nodes but underrepresent cross-zone connectivity slightly.

Node matching between human-labeled and VLM-generated graphs shows that the majority of human-labeled nodes are successfully matched to their VLM counterparts, with unmatched rates around or below 10\%. Higher unmatched rates are observed for functionally peripheral categories such as auditorium (cat8: 21.1\%) and courtyard/outdoor spaces (cat0: 31.4\%), where spatial boundaries are less well defined in floor plan images and these spaces appear less frequently in library plans. The unmatched ratio for VLM nodes is higher, but this is primarily a consequence of VLM-generated graphs containing more nodes overall rather than an indication of inconsistency. For instance, the unmatched ratios of cat3 (reading seats), cat6 (self-study), and cat13 (others) exceed 20\%, mainly because human annotators tend to merge such zones into single nodes, whereas the VLM resolves them separately.

Edge co-occurrence matrices show broadly similar functional co-occurrence patterns across both graph types. Both the node function alignment matrix and the edge type alignment matrix exhibit a clear diagonal structure, confirming that the matched node are actually consistent in functions and locations. These results support the validity of VLM-generated graphs as a structurally faithful approximation of human-labeled representations and justify their use in subsequent multi-granularity analyses.

\subsubsection{Multi-granularity graph comparison}
\label{sec:results-fid-multigran}

\begin{figure}[!htbp]
\centering
\includegraphics[width=\linewidth,height=0.7\textheight,keepaspectratio]{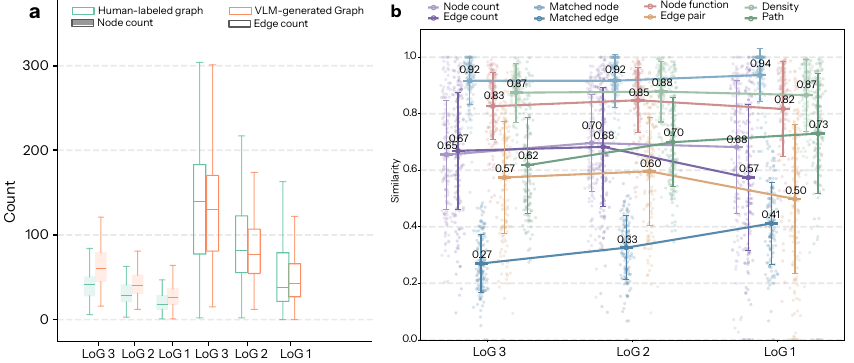}
\caption{Structural comparison of human-labeled and VLM-generated graphs across three LoG levels. (a) Node and edge count distributions; (b) Similarity metrics between matched graph pairs.}
\label{fig:lod-similarity}
\end{figure}

Figure~\ref{fig:lod-similarity} extends the fidelity evaluation to all three LoG levels. At every level, VLM-generated graphs contain more nodes than their human-labeled counterparts while edge counts remain broadly comparable. The node count gap narrows at LoG~2 and further at LoG~1, consistent with coarsening reducing the influence of initial scale differences.

Among the similarity metrics, matched node ratio, density similarity and node function distribution similarity are consistently the highest across all granularity levels: despite differences in node count, the overall functional composition is well preserved in VLM-generated graphs. Edge count similarity and edge function pair similarity decline further at LoG~1. This indicates that while individual functional zones are comparably represented, the specific spatial relations between functional categories diverge, likely reflecting the VLM's weaker capacity for recognizing complex cross-zone connectivity. 

The matched edge ratio is the lowest, but this is reasonable considering it requires both endpoints to be matched, and the goal of VLM-generated graphs is not to reproduce the exact same nodes as the human-labeled graphs. Given that the matched node ratio exceeds 0.92 and node function, density, and path similarities all remain high, VLM-generated graphs faithfully reproduce the overall scale, functional composition, and topological reachability of human-labeled graphs. More importantly, the matched edge ratio rises across granularity levels, indicating that coarsening absorbs local disagreements into shared super-nodes and exposes the macro-level connectivity structure. Taken together, these results show that VLM-generated graphs can serve as functionally valid proxies for human-labeled graphs in downstream design analysis tasks, though VLM LoG~1 (coarse-grained) graphs may introduce minor distortions in local inter-functional connectivity due to the compounding effect of coarsening.

\subsection{Downstream task effectiveness with feature importance analysis}
\label{sec:results-effectiveness}

This subsection evaluates the multi-granularity graphs on task performance with computational complexity across graph sources and granularity levels. 

\subsubsection{Task performance and computational complexity across granularity levels}
\label{sec:results-perf}

Table~\ref{tab:main-results} reports task performance across graph sources and granularity levels for both downstream tasks.

\begin{table}[!htbp]
\centering
\caption{Multi-task performance of graph representations across three LoG levels.}
\label{tab:main-results}
\footnotesize
\setlength{\tabcolsep}{4pt}
\renewcommand{\arraystretch}{1.15}
\begin{threeparttable}
\begin{tabular}{lllccc}
\toprule
\textbf{Task} & \textbf{Graph type} & \textbf{Model} & \textbf{LoG~3 (fine)} & \textbf{LoG~2 (meso)} & \textbf{LoG~1 (coarse)} \\
\midrule
\multirow{6}{*}{\shortstack[l]{Layout quality\\evaluation (Spearman's~$\rho$)}}
  & \multirow{3}{*}{Human} & XGBoost  & 0.508 & 0.323 & 0.439 \\
  &                        & Ridge    & 0.480 & 0.428 & \textbf{0.610*} \\
  &                        & SVR-RBF  & 0.435 & 0.317 & 0.542 \\
\cmidrule(lr){2-6}
  & \multirow{3}{*}{VLM}   & XGBoost  & 0.151 & 0.050 & 0.135 \\
  &                        & Ridge    & 0.510 & 0.214 & $-0.026$ \\
  &                        & SVR-RBF  & 0.449 & 0.109 & $-0.017$ \\
\midrule
\multirow{8}{*}{\shortstack[l]{Zone function\\prediction (Macro F1)}}
  & \multirow{4}{*}{Human} & GINE      & 0.660 & \textbf{0.647*} & / \\
  &                        & GCN       & 0.555 & 0.583           & / \\
  &                        & GraphSAGE & \textbf{0.662} & 0.618           & / \\
  &                        & GAT       & 0.548 & 0.487           & / \\
\cmidrule(lr){2-6}
  & \multirow{4}{*}{VLM}   & GINE      & 0.573 & 0.554           & / \\
  &                        & GCN       & 0.574 & 0.492           & / \\
  &                        & GraphSAGE & 0.582 & 0.560           & / \\
  &                        & GAT       & 0.516 & 0.555           & / \\
\bottomrule
\end{tabular}
\begin{tablenotes}[flushleft]\footnotesize
\item \textbf{Bold} $=$ best performance; \textbf{Bold $+$ *} $=$ recommended under the performance--complexity trade-off. Nodes at the coarse level are multi-functional, making node prediction not applicable.
\end{tablenotes}
\end{threeparttable}
\end{table}

\begin{table}[!htbp]
\centering
\caption{Graph scale and computational complexity across three LoG levels.}
\label{tab:complexity}
\footnotesize
\setlength{\tabcolsep}{6pt}
\renewcommand{\arraystretch}{1.15}
\begin{threeparttable}
\begin{tabular}{llccc}
\toprule
\textbf{Graph type} & \textbf{Metric} & \textbf{LoG~3 (fine)} & \textbf{LoG~2 (meso)} & \textbf{LoG~1 (coarse)} \\
\midrule
\multirow{3}{*}{Human} & scale              & 41n,\,146e & 32n,\,95e & 22n,\,58e \\
                       & $r_{\ell}^{\feat}$ & 1.00       & 0.42      & 0.16      \\
                       & $r_{\ell}$         & 1.00       & 0.65      & 0.40      \\
\cmidrule(lr){1-5}
\multirow{3}{*}{VLM}   & scale              & 62n,\,130e & 43n,\,85e & 30n,\,54e \\
                       & $r_{\ell}^{\feat}$ & 1.00       & 0.42      & 0.16      \\
                       & $r_{\ell}$         & 1.00       & 0.65      & 0.42      \\
\bottomrule
\end{tabular}
\begin{tablenotes}[flushleft]\footnotesize
\item Scale rows show mean node ($n$) and edge ($e$) counts per graph.
\item $r_{\ell}^{\feat} = (\card{V_{\ell}} \cdot \card{E_{\ell}}) / (\card{V_{\fine}} \cdot \card{E_{\fine}})$: relative feature extraction complexity.
\item $r_{\ell} = \card{E_{\ell}} / \card{E_{\fine}}$: relative GNN complexity.
\end{tablenotes}
\end{threeparttable}
\end{table}
For layout quality evaluation, the strongest performance was observed at LoG~1 (Ridge regression, Spearman $\rho = 0.610$). LoG~3 human-labeled graphs also achieved competitive performance ($\rho = 0.508$). At LoG~3, performance was virtually identical across graph sources (human-labeled: $\rho = 0.508$; VLM: $\rho = 0.510$), demonstrating that VLM-generated graphs can be a viable substitute for human-labeled graphs in quality evaluation.    

For zone function prediction, cat4 (Lounge) and cat6 (Self-study) were consistently near-unpredictable in the five-class setting (see Figure~\ref{fig:class-heatmap}), likely because these functions are typically placed residually in library layouts rather than positioned according to strong topological constraints. When restricting prediction to the other three topologically distinct categories, Collections (cat2), Reading Seats (cat3), and Group Study (cat5), LoG~3 and LoG~2 graphs performed comparably\footnote{Macro-F1 scores under the 5-class setting are reported in Appendix Table~\ref{tab:appendix-results}. The granularity-task pattern remains consistent.}. LoG~3 graphs preserve the full local connectivity of individual furniture groups, while LoG~2 graphs retain comparable discriminability by consolidating co-functional clusters while preserving inter-zone connectivity.

Specifically, the performance values should be interpreted in light of task difficulty. Layout quality evaluation requires predicting holistic expert aesthetic judgements. Zone function prediction masks all 5-class user-oriented functional nodes simultaneously, which is a substantially more demanding setting than partial label propagation. Competitive Spearman correlations and macro-F1 scores are achieved under these constraints, suggesting that the graph representations capture genuinely informative spatial signals.

Granularity selection also entails a computational trade-off (Table~\ref{tab:complexity}). LoG~2 graphs reduce GNN complexity ($r_{\ell}$) by approximately 35\% and feature extraction cost ($r_{\ell}^{\feat}$) to 42\% of the LoG~3 baseline, while maintaining competitive zone function prediction performance. LoG~1 graphs achieve a further reduction to 40\% GNN complexity ($r_{\ell}$) and 16\% feature extraction cost ($r_{\ell}^{\feat}$) and simultaneously deliver the strongest layout quality regression performance. These results suggest: (i)~LoG~1 graphs are preferred for graph-level tasks such as layout quality evaluation, being both computationally cheaper and more predictively effective; (ii)~LoG~2 graphs offer the best complexity--discriminability balance for node-level tasks such as zone function prediction; (iii)~LoG~3 graphs remain appropriate when maximum local topological fidelity is required and computational cost is unconstrained.

\subsubsection{Architectural knowledge from feature importance}
\label{sec:results-feat-importance}

We visualized top-predicted examples for both tasks to illustrate the practical usability of the proposed LoGs in real application scenarios. Feature importance analyses further reveal interpretable architectural knowledge encoded in the multi-granularity graph representations.

\begin{figure}[!htbp]
\centering
\includegraphics[width=\linewidth]{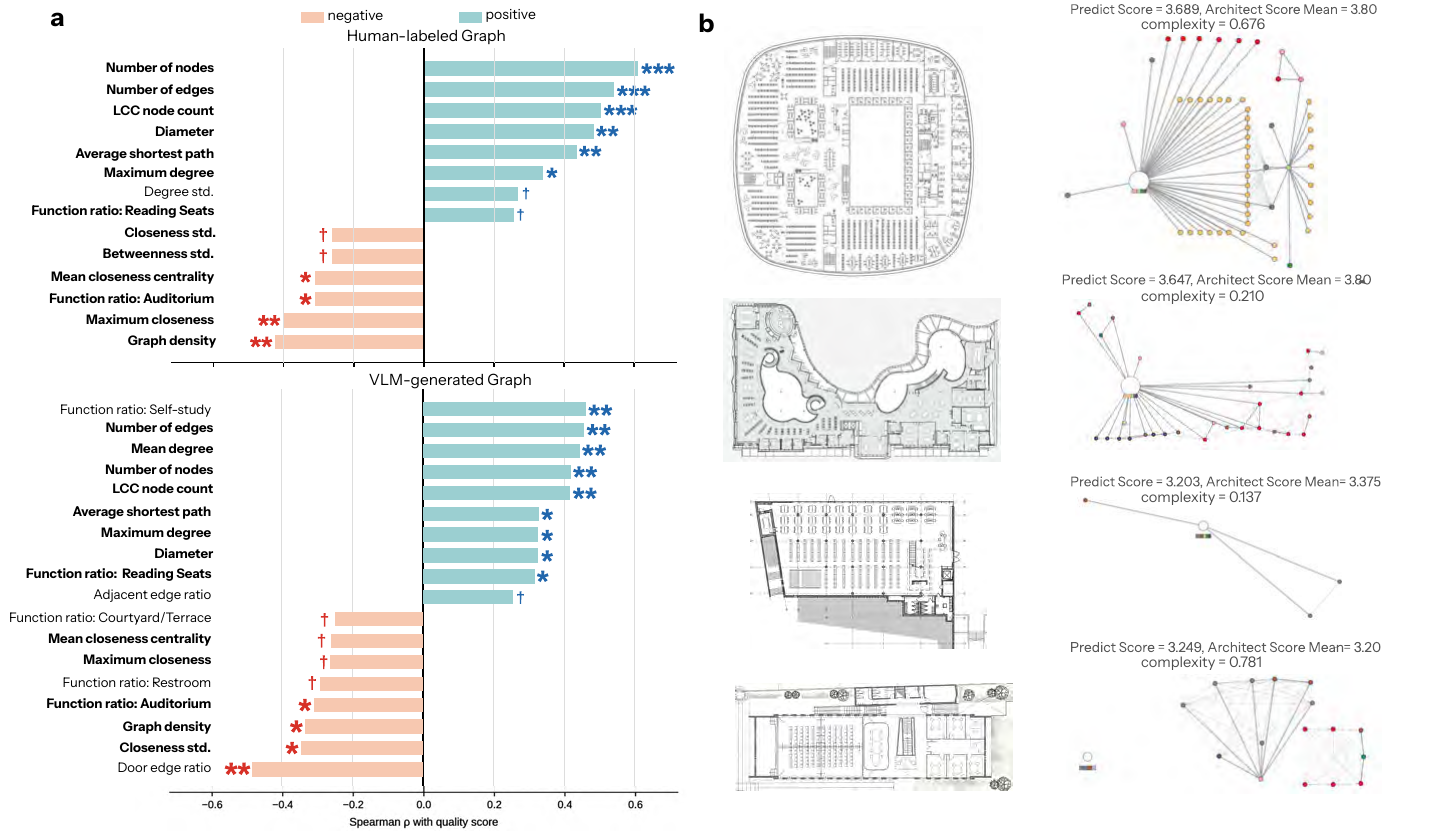}
\caption{Layout quality evaluation. (a)~Feature importance for layout quality evaluation. $\dagger\,p<0.1$, ${}^*\,p<0.05$, ${}^{**}\,p<0.01$, ${}^{***}\,p<0.001$. (b) LoG~1 graphs of top- (top two rows) and bottom-quality cases (bottom two rows). }
\label{fig:feat-importance}
\end{figure}

For layout quality evaluation, 13 of the 42 features show significant correlations with consistent directions across both human-labeled and VLM-generated graphs (Figure~\ref{fig:feat-importance}a), further demonstrating the effectiveness of VLM-generated graphs. Among the shared significant features, graph scale dominates. Beyond scale, diameter and average shortest path also show significant positive associations, while graph density, maximum closeness, and mean closeness centrality are negatively correlated with quality scores. Collectively, these results suggest that larger, more spatially extended rather than compact library layouts tend to be considered higher quality by experienced architects. Maximum degree and degree standard deviation contribute positively, indicating that higher layout quality is associated with the presence of high-connectivity functional hubs and greater heterogeneity in spatial connectivity. This implies that well-rated library layouts tend to provide diverse zones that balance openness and privacy. Regarding functional composition, a higher proportion of reading seats is positively associated with quality, while a higher proportion of auditorium space shows a significant negative association, consistent with its peripheral and space-consuming character in library configurations. In the presented LoG~1 examples from the top-10 and bottom-10 cases, clear differences are observed in graph scale, the number of nodes in the largest connected component (LCC), and the presence of high-degree multifunctional nodes (Figure \ref{fig:feat-importance}b). These observations are consistent with the feature importance analysis and further demonstrate the interpretability and practical effectiveness of the LoG~1 representation.

\begin{figure}[!htbp]
\centering
\includegraphics[width=\linewidth]{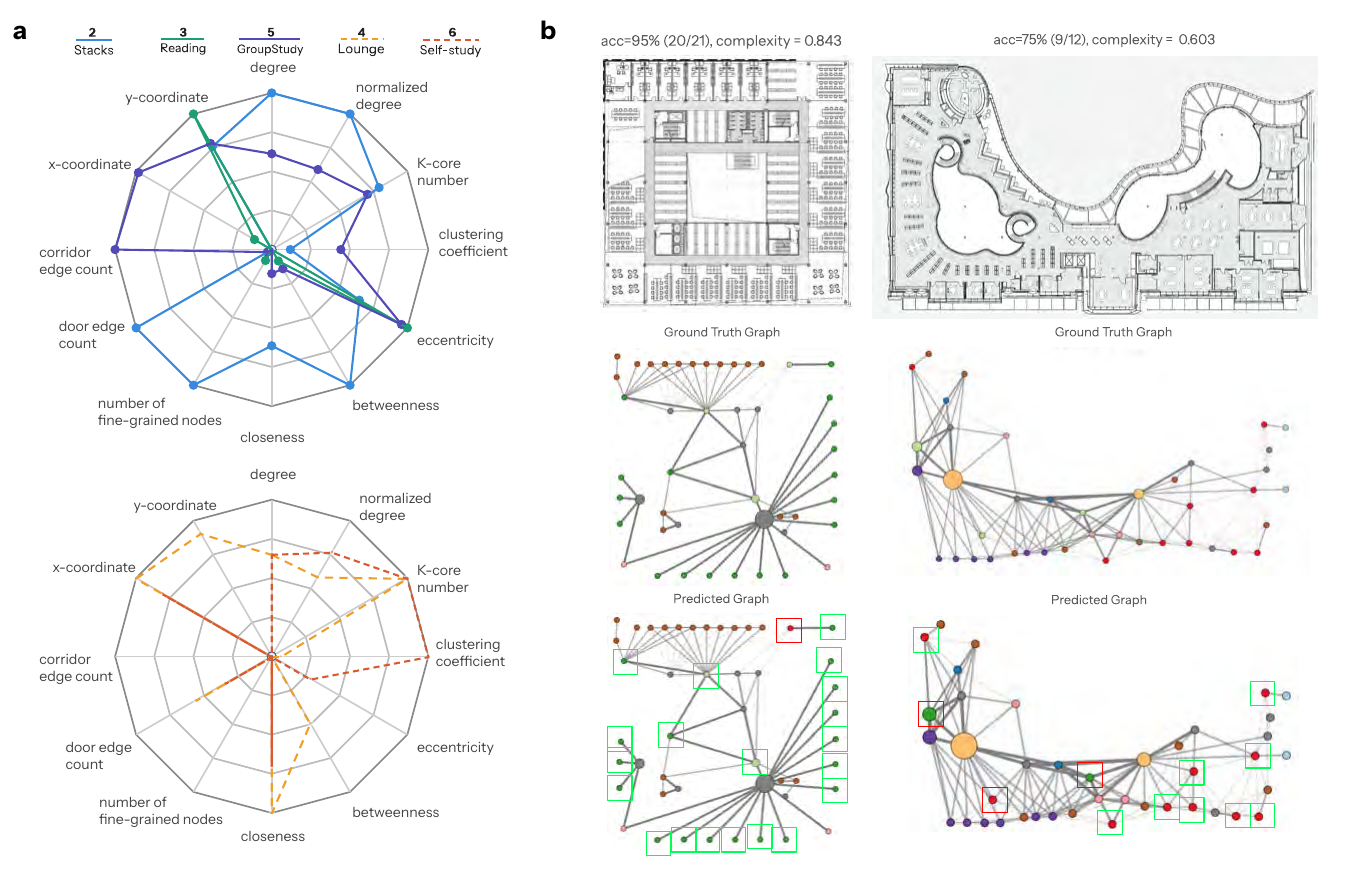}
\caption{Zone function prediction. (a)~Node feature profiles for zone function prediction. (b) LoG~2 graphs of top-accuracy cases.} 
\label{fig:zone}
\end{figure}

For zone function prediction, the node feature profiles reveal distinct topological signatures for each functional category (Figure \ref{fig:zone}a) , which can inversely inform their arrangement and design. Collections (cat2) is characterized by high degree, high betweenness, and high K-core number, confirming its topologically central position as a heavily connected hub within the layout. Reading Seats (cat3) show high eccentricity and $y$-coordinate values, indicating a peripheral spatial position but serving as a structurally accessible endpoint reachable from multiple zones, which reflects the balance of privacy and accessibility that design should meet. Group Study (cat5) presents high eccentricity, door edge count, and $x$-coordinate values, suggesting it typically occupies a laterally peripheral and enclosed position with explicit door connections. Lounge (cat4) and Self-study (cat6) exhibit less differentiated profiles overall, reflecting the diversity of their spatial arrangements across cases. Nevertheless, subtle distinctions are observable: Lounge shows relatively high closeness centrality despite low degree, suggesting it should be arranged at globally accessible locations such as entrance areas or atriums without being heavily connected topologically. Self-study is characterized by high K-core number and clustering coefficient, indicating it is embedded within locally dense, mutually connected clusters of nodes. In the presented top-accuracy LoG~2 examples, the model correctly predicts the majority of bookshelves, reading areas, and discussion zones (Figure \ref{fig:zone}b, green boxes). Misclassifications are limited to enclosed spaces that are incorrectly identified as discussion zones, as well as highly central bookshelf areas that are misclassified as reading zones (Figure \ref{fig:zone}b,  red boxes).

\section{Discussion}
\label{sec:discussion}

\subsection{Significance and implications}
\label{sec:significance}

The significance of this work lies in automatically converting public building layouts into computationally operable representations that can support a wide range of downstream design and engineering tasks. \begin{itemize}
    \item Theoretically, the proposed LoGs extends the multi-granularity principle established in urban informatics into building layouts, bridging the long-standing gap between visually rich but computationally inaccessible floor plans and data-driven design workflows. 
    \item Methodologically, the proposed VLM-based pipeline eases the manual annotation bottleneck that has long constrained graph-based architectural research. This makes the large-scale construction of layout graphs feasible for public buildings. 
    \item Empirically, the granularity--task alignment revealed in this paper shows that no single graph granularity dominates across analytical tasks, positioning granularity as an explicit design decision rather than a fixed property of the representation. This paper also supports design decisions for academic libraries by providing explicit architectural knowledge in terms of design strategies that can enhance spatial quality.

\end{itemize}

More importantly, the availability of multi-granularity graph representations opens up a range of practical applications. For architects, layout graphs enable rapid case retrieval and comparison, supporting precedent-based design and early-stage decision-making. For urban planners, different graph granularities allow targeted analysis of building footprint, functional allocation, and user experience within a unified framework, thereby enhancing collaboration between urban planners and architects. For engineers, graph representations facilitate the integration of design-stage layouts with BIM generation, enrichment and validation.

Overall, this study advances design informatics by transforming floor plans from static visual artifacts into scalable, reusable, and computation-ready knowledge, with direct implications for design analysis, decision-making, and intelligent design systems.

\subsection{Key findings}
\label{sec:findings}

The paper yields three sets of findings that can inspire broader insights.

\textbf{Building layout reasoning capacities of VLM}. Our findings show that current VLMs are better calibrated for local functional recognition than for relational spatial reasoning. Functional zone identification is a local perceptual task: visually distinctive furniture configurations provide reliable cues that VLMs can exploit. Spatial connectivity, by contrast, is relational and often implicit. Reasoning about relations is not merely a matter of retrieving stored knowledge, but requires integrating non-local context and interpreting subtle graphic conventions that are not well supported by the local attention mechanisms \citep{raghu2021vision,radford2021learning} of transformer-based vision models \citep{gemini2025gemini,openai2024gpt4}. Although slight divergences are observed, considering that a graph is an abstraction of a real-world object and that there is no single absolutely correct representation, these differences from the human-labeled graphs do not theoretically undermine the validity of the VLM-generated representations, especially given that they have been proven effective in downstream tasks. Improving VLM-based graph extraction will therefore require targeting relational spatial reasoning specifically, whether through explicit topological consistency constraints during graph construction or by incorporating three-dimensional spatial understanding \citep{li2025efficient,Li2026UrbanNatureHappiness} to resolve ambiguities inherent in two-dimensional plan projections.

\textbf{Significance of multi-granularity representation.} The results reveal a granularity--task interaction: coarse- and meso-grained graphs achieve optimal performance on different tasks, indicating that optimal graph granularity is task-dependent rather than universally determined by spatial resolution. Fine-grained graphs, though associated with relatively less information loss, can introduce local redundancies. In essence, coarsening acts as a feature selection mechanism: it discards locally redundant detail and concentrates the most task-relevant spatial signals. Although fine-grained graphs achieve comparable performance across the tasks evaluated in this study, the substantial differences in computational complexity across granularity levels make multi-granularity graph representations practically necessary.

\textbf{Toward human--VLM collaborative spatial cognition.} Human-labeled graphs excel at relational spatial reasoning by integrating architectural domain knowledge and three-dimensional reasoning that cannot be directly read from a two-dimensional projection. VLM-generated graphs, by contrast, demonstrate reliable functional recognition and consistent compositional representation. A collaborative workflow that combines VLM-based functional recognition with human-based relational reasoning could capture the strengths of both in broad tasks of design cognition. More broadly, the finding that VLM-generated graphs achieve prediction performance comparable to human-labeled graphs at the fine granularity suggests that architectural knowledge may be robustly encoded across multiple representational formats. 

\subsection{Limitations and future directions}
\label{sec:discussion-limits}

Despite these contributions, this study has several limitations. 
\begin{itemize}
    \item First, the capability of VLMs to identify additional attributes such as areas has not been evaluated. The absence of such attributes constrains coarse-level semantic fidelity, and future work should explore geometric attribute estimation, such as zero-shot bounding box generation or instance segmentation for complex building layouts. 
    \item Second, the feature sets used in downstream tasks are constructed based on architects' knowledge and may not capture all spatially relevant information. Conclusions are therefore valid within the scope of the current feature sets, and future work should extend the evaluation to a broader range of downstream tasks, including generative tasks such as layout synthesis and optimization, as well as integration with broader AEC workflows such as BIM enrichment, semantic interoperability, and computational compliance checking.
    \item Third, the dataset comprises 147 academic library floor plans, and the generalizability to other building typologies remains to be verified. Extending the LoGs  to other public building types would further establish multi-granularity architectural graph representation as a foundational tool for computational design automation.
\end{itemize}

\section{Conclusion}
\label{sec:conclusion}

This study proposed an automatic multi-granularity graph representation for complex public building layouts, using academic libraries as a case study. We introduced the Level-of-Graphs (LoGs) that represents building layouts at different level of details. We also developed a VLM-based pipeline that constructs such representations from floor plan images without manual annotation. Evaluation metrics were organized to assess the structural fidelity and computational complexity of the constructed graphs.

Experimental results demonstrate that the pipeline produces structurally faithful graphs: VLM-generated graphs generally match human-labeled graphs in functional composition and global configuration (matched node ratio >= 92\%; 509.3 s per floor plan for three-LoG graph generation), though they exhibit limitations in capturing relational cross-zone connectivity. Across two representative downstream tasks, optimal graph granularity was found to be task-dependent: meso-grained (LoG~2) graphs best support node-level zone function prediction (Macro-F1 = 0.647, at 65\% of fine-grained complexity), while coarse-grained (LoG~1) graphs best support graph-level layout quality evaluation (Spearman's~$\rho$ = 0.610, at 16\% of fine-grained complexity). These findings establish granularity--task alignment as a foundation for graph-based architectural analysis.

Beyond the empirical findings, broader observations emerge. First, the asymmetry between VLM-generated and human-labeled graphs suggests that architectural spatial cognition can be decomposable into complementary components. This inspires a human--VLM collaborative workflow in which VLM-based element recognition and human-based relational inference are combined as substitutable modules. Second, coarsening can act as a feature selection mechanism, discarding locally redundant detail to concentrate task-relevant spatial signals. This reframes granularity as an analytical choice that decides what spatial knowledge a model can access.

This study has several limitations and can be addressed in future work. The capability of VLMs to infer geometric attributes was not evaluated. Incorporating geometric attribute estimation and three-dimensional spatial understanding has the potential to address the VLM's relational reasoning gap. The feature sets used in downstream tasks are derived from domain knowledge. Testing the multi-granularity representations in generative tasks and integrating them with BIM enrichment would further validate their utility across AEC workflows. The dataset consists of 147 academic library floor plans. Extending the automatic, annotation-free pipeline to other building typologies would make it suitable for integration into broader architectural decision-support processes. By linking automatic graph extraction with multi-granularity representation and downstream task applications, this work contributes to design informatics a scalable, annotation-efficient pathway for formalizing complex architectural information into task-adaptive structured representations.

\section*{Acknowledgements}
We thank Chaoyi Huang for the discussion on methodology and conceptualization, and all 12 experienced architects for their evaluation of the layout quality.

\section*{CRediT authorship contribution statement}
\textbf{Song Guo:} Writing -- review \& editing, Writing -- original draft, Visualization, Validation, Software, Methodology, Investigation, Formal analysis, Data curation, Conceptualization.
\textbf{Zhuoshi Chen:} Writing -- original draft, Visualization, Software, Formal analysis, Methodology.
\textbf{Maosu Li:} Writing -- review \& editing, Validation, Methodology, Conceptualization.
\textbf{Weimin Zhuang:} Writing -- review \& editing, Data curation.
\section*{Declaration of generative AI and AI-assisted technologies in the manuscript preparation process}
During the preparation of this work the authors used Claude in order to assist with table formatting, and language editing of the manuscript. After using this tool/service, the authors reviewed and edited the content as needed and take full responsibility for the content of the published article.

\bibliographystyle{elsarticle-num-isbn}
\bibliography{refs}
\

\appendix
\renewcommand{\thefigure}{\thesection.\arabic{figure}}
\renewcommand{\thetable}{\thesection.\arabic{table}}
\makeatletter
\@addtoreset{figure}{section}
\@addtoreset{table}{section}
\let\origthesection\thesection
\renewcommand{\thefigure}{\@Alph\c@section.\arabic{figure}}
\renewcommand{\thetable}{\@Alph\c@section.\arabic{table}}
\makeatother

\section{VLM graph construction experiments}
\label{app:vlm}
This appendix provides the prompts used for VLM-based graph construction, together with a side-by-side comparison of four SOTA VLMs on node identification that informed the choice of Gemini~3~Pro as the primary model (Figure~\ref{fig:appendix-vlm-compare}).

\subsection{Node identification prompt}
\label{app:node-prompt}
\begin{promptquote}
\textbf{Role:} You are an expert in architectural floorplan semantic graph annotation.\par\vspace{2pt}\noindent
\textbf{Goal:} Extract ALL fine-grained functional sub-space nodes from the FULL floorplan image.\par\vspace{2pt}\noindent
Output ONLY plain text. NO explanations. NO markdown.\par\vspace{6pt}\noindent
\textbf{[Global Analysis First - MUST]}\par
Before outputting anything:\par
- Scan the ENTIRE floorplan.\par
- Identify ALL functional furniture clusters and enclosed rooms.\par
- Then filter by region constraint below.\par\vspace{6pt}\noindent
\textbf{[Coordinate System]}\par
- Top-left is (0,0), bottom-right is (1000,1000).\par
- Output normalized coordinates in this 0..1000 space.\par\vspace{6pt}\noindent
\textbf{[REGION FILTER - MUST FOLLOW]}\par
You MUST output ONLY nodes whose CENTER (X,Y) satisfies:\par
X in [\{x0\}, \{x1\}] AND Y in [\{y0\}, \{y1\}].\par\vspace{2pt}\noindent
\textit{Important:}\par
- You are seeing the FULL floorplan. Use the FULL context to infer the true activity center.\par\vspace{6pt}\noindent
\textbf{[Node Definition - MUST FOLLOW]}\par
- One node = one distinct functional sub-space (not necessarily a walled room).\par
- Spaces are inferred from furniture grouping, functional zoning, labels, partitions, and spatial layout.\par
- A node should represent a coherent functional area.\par
- Repeated or continuous identical furniture patterns within one coherent space (e.g., rows of shelves or tables) should be treated as a single node (or a small number of nodes) rather than one node per row.\par\vspace{6pt}\noindent
\textbf{[Empty Enclosed Rooms - MUST]}\par
- You MUST detect enclosed rooms defined by continuous walls, even if no door symbol is explicitly drawn.\par
- A room may be considered enclosed if it is spatially bounded by walls and has a clear opening or access connection to adjacent space (even without a drawn door leaf).\par
- Such enclosed empty rooms MUST be output as nodes.\par
- For these rooms, place the point at the geometric center of the enclosed area.\par
- Unless clear evidence suggests another category, classify empty enclosed rooms as (13) Other.\par
- Do NOT classify enclosed empty rooms as (12) Circulation.\par\vspace{6pt}\noindent
\textbf{[Distinguish Carefully:]}\par
\textbf{(7) Classroom:} enclosed room; many desks/chairs; ALL desks face the same direction (teaching orientation).\par
\textbf{(6) Self-study:} single-person desks OR long desks facing wall/atrium; layout suggests individual focus.\par
\textbf{(5) Group Study:} shared tables for discussion; not strictly single-direction; can be enclosed or open.\par
\textbf{(3) Reading Commons:} many repeated 4-person / 6-person tables; large general reading hall pattern.\par
\textbf{(1) Reception / Information:} clear counter near entrance; often linear or semi-circular counter shape; staff-facing orientation.\par
\textbf{(9) Electronic Reading / Specialized:} computer desks, monitor symbols, equipment; or special facilities (studio, workshop, exhibition).\par
\textbf{(12) Circulation:} wall-defined corridor OR negative space between furniture clusters; do NOT label furniture cluster as 12.\par
\textbf{(13) Other:} when furniture pattern cannot support any classification (e.g., empty room without any furniture).\par\vspace{6pt}\noindent
\textbf{[CategoryID MUST be an INTEGER from 0 to 13]}\par

\textbf{[Hard Rules]}\par
- Do NOT output category names. Output ONLY the integer CategoryID.\par
- Use ASCII vertical bar: \textbar\par
- IDs must start at 0 and be consecutive WITHIN THIS REGION RESPONSE.\par\vspace{6pt}\noindent
\textbf{[Point Placement]}\par
- Place point at the activity center of the sub-space.\par
- Avoid placing points on walls or text labels.\par\vspace{6pt}\noindent
\textbf{[Output Format]}\par
Each line: ID\textbar CategoryID\textbar X\textbar Y
\end{promptquote}
 
\subsection{Edge inference prompt}
\label{app:edge-prompt}
 
\begin{promptquote}
\textbf{Inputs:} You are given TWO images:\par
- Image A: reference map with red dots (node centers) and yellow node IDs.\par
- Image B: the original floorplan image (same scale as Image A).\par\vspace{6pt}\noindent
\textbf{Task:} For the single CENTER node \{center\_id\}, list ONLY its DIRECTLY ADJACENT neighbors.\par\vspace{2pt}\noindent
Output ONLY plain text. NO explanations. NO markdown.\par\vspace{6pt}\noindent
\textbf{[How to use images]}\par
- Use Image A to identify node IDs precisely.\par
- Use Image B to understand walls, doors, partitions, and functional boundaries.\par
- Cross-check both images before deciding neighbors.\par\vspace{6pt}\noindent
\textbf{[Direct Adjacency - LOCAL, PRACTICAL]}\par
A neighbor is DIRECTLY ADJACENT if:\par
1) The two sub-spaces are immediate local neighbors in the plan, AND\par
2) You can pass from one to the other with only a very short transition, as long as that transition is NOT itself a distinct functional sub-space.\par\vspace{2pt}\noindent
\textit{Important constraints:}\par
- Do NOT create long-range edges across large halls (no ``across the room'' connections).\par
- Do NOT connect two nodes if another functional node/zone clearly lies in-between them.\par
- Being in the same open-plan room is NOT enough; adjacency must be local and immediate.\par\vspace{6pt}\noindent
\textbf{[Edge Types]}\par
- H = ``door / hard connection'': explicit opening/door across a wall/partition.\par
- S = ``open / soft connection'': immediate open adjacency (no separating wall/partition).\par\vspace{6pt}\noindent
\textbf{[Output Format]}\par
Exactly ONE line:\par
CenterID\textbar NeighborID1:Type,NeighborID2:Type,...\par
Where Type is H or S.\par
If no neighbors, output: CenterID\textbar
\end{promptquote}

\begin{center}
\includegraphics[width=\linewidth]{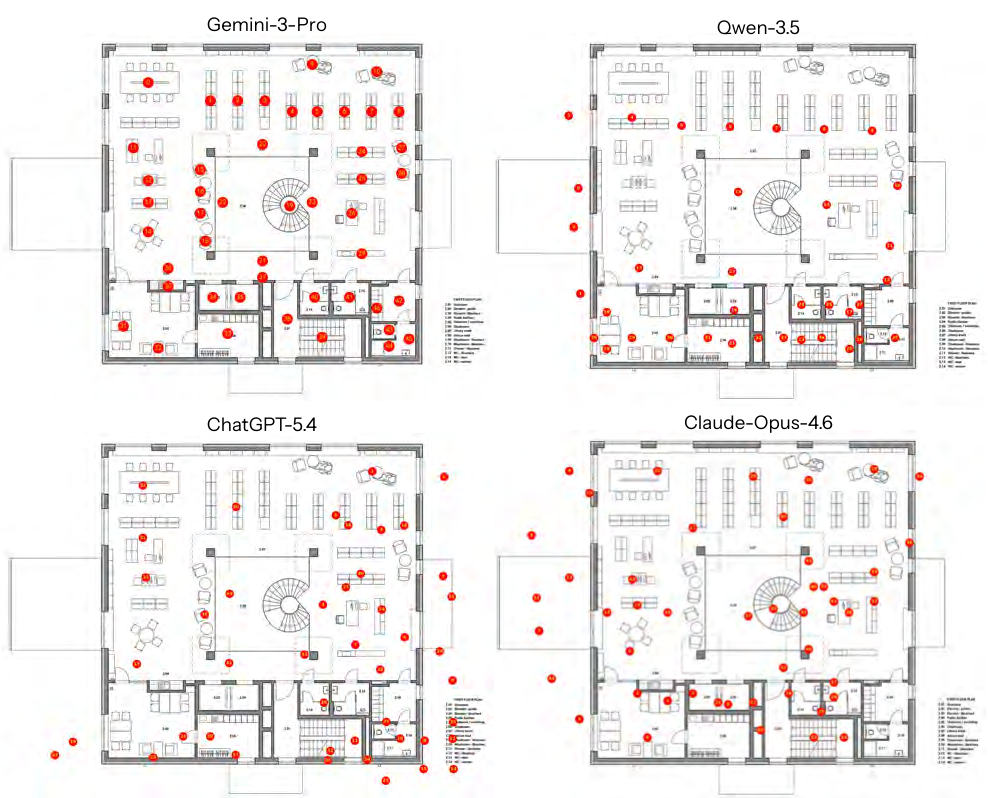}
\captionof{figure}{Node identification experiments across four SOTA VLMs on a representative floor plan.}
\label{fig:appendix-vlm-compare}
\end{center}

\section{Downstream experiments}
\label{app:downstream}
Table~\ref{tab:graph-features} and Table~\ref{tab:node-features} presents graph-level features and node-level features used in the downstream tasks separately.
{\footnotesize
\setlength{\tabcolsep}{4pt}
\renewcommand{\arraystretch}{1.3}
\begin{longtable}{@{}>{\raggedright\arraybackslash}p{2.2cm} >{\raggedright\arraybackslash}p{2.8cm} >{\raggedright\arraybackslash}p{8.0cm} >{\raggedright\arraybackslash}p{1.8cm}@{}}
\caption{Graph-level feature set for multi-granularity layout graphs.}\label{tab:graph-features}\\
\toprule
\textbf{Feature Group} & \textbf{Feature Name} & \textbf{Definition / Formula} & \textbf{Notation} \\
\midrule
\endfirsthead
\toprule
\textbf{Feature Group} & \textbf{Feature Name} & \textbf{Definition / Formula} & \textbf{Notation} \\
\midrule
\endhead
\bottomrule
\multicolumn{4}{r}{\textit{(continued on next page)}}\\
\endfoot
\bottomrule
\endlastfoot
Topology
 & Number of nodes       & Total count of nodes in graph $G = (V, E)$ & $\card{V}$ \\
 & Number of edges       & Total count of edges in graph $G = (V, E)$ & $\card{E}$ \\
 & Graph density         & $d = \card{E} \,/\, \bigl(\card{V}(\card{V}-1)/2\bigr)$ & $d$ \\
 & Connected components  & Count of maximal connected subgraphs in $G$ & $c$ \\
 & Degree assortativity  & Pearson correlation of degrees across connected node pairs \citep{newman2003mixing}; $r \in [-1, 1]$ & $r$ \\
 & Avg.\ clustering coef.
   & $\bar{C} = \frac{1}{\card{V}} \sum_i \frac{2 t_i}{k_i(k_i-1)}$, where $t_i$ = triangles at node $i$, $k_i$ = degree of node $i$ & $\bar{C}$ \\
 & Transitivity          & $T = 3 \times (\text{closed triangles}) / (\text{open} + \text{closed triangles})$ & $T$ \\
\midrule
Degree
 & Mean degree           & $\mu_k = \frac{1}{\card{V}} \sum_i k_i$ where $k_i$ = degree of node $i$ & $\mu_k$ \\
 & Degree std.\ dev.     & Standard deviation of node degrees across $G$ & $\sigma_k$ \\
 & Maximum degree        & $\max\{k_i : i \in V\}$ & $k_{\max}$ \\
 & Minimum degree        & $\min\{k_i : i \in V\}$ & $k_{\min}$ \\
\midrule
Path
 & LCC node count        & Number of nodes in the largest connected component (LCC) of $G$ & $\card{V_L}$ \\
 & Avg.\ shortest path (LCC)
   & $\bar{\ell} = \frac{1}{\card{V_L}(\card{V_L}-1)} \sum_{i \neq j} d(i,j)$ on LCC; $d(i,j)$ = shortest path length & $\bar{\ell}$ \\
 & Diameter (LCC)        & $D = \max_{i,j \in V_L} d(i,j)$, longest shortest path in LCC & $D$ \\
\midrule
Centrality
 & Mean betweenness      & $C^b(v) = \sum_{s \neq v \neq t} \sigma(s,t \mid v) / \sigma(s,t)$; $\mu^b$ = mean over all $v \in V$ ($\sigma$ = number of shortest paths) & $\mu^b$ \\
 & Betweenness std.\ dev.& Standard deviation of $C^b(v)$ over all nodes & $\sigma^b$ \\
 & Maximum betweenness   & $\max\{C^b(v) : v \in V\}$ & $C^b_{\max}$ \\
 & Mean closeness        & $C^c(v) = (\card{V}-1) / \sum_{u \neq v} d(u,v)$; $\mu^c$ = mean over all $v$ & $\mu^c$ \\
 & Closeness std.\ dev.  & Standard deviation of $C^c(v)$ over all nodes & $\sigma^c$ \\
 & Maximum closeness     & $\max\{C^c(v) : v \in V\}$ & $C^c_{\max}$ \\
\midrule
Function
 & Function class ratio
   & $f_n = \card{\{v : \text{func}(v) = n\}} / \card{V}$, proportion of nodes assigned to function class $n$. 
   & $f_n$ \\
 & Function entropy      & $H_f = - \sum_n f_n \log f_n$, Shannon entropy over function class distribution & $H_f$ \\
\midrule
Composition
 & Members mean          & $\mu_m = \frac{1}{\card{V}} \sum_{v \in V} n_{\text{members},v}$, where $n_{\text{members},v}$ = number of fine-level nodes merged into $v$ & $\mu_m$ \\
 & Members std.\ dev.    & $\sigma_m = \mathrm{std}\{n_{\text{members},v} : v \in V\}$ & $\sigma_m$ \\
 & Maximum members       & $\max\{n_{\text{members},v} : v \in V\}$ & $k_{m,\max}$ \\
 & Composition entropy mean
   & $\bar{H}_{\text{comp}} = \frac{1}{\card{V}} \sum_{v \in V} H(p_v)$, where $H(p_v) = -\sum_c p_{v,c} \log p_{v,c}$; $p_{v,c}$ = proportion of function $c$ among members of $v$. 
   & $\bar{H}_{\text{comp}}$ \\
 & Multifunctional node ratio
   & $r_{\text{multi}} = \card{\{v : \card{\text{func\_count\_dict}_v} > 1\}} / \card{V}$; fraction of nodes containing more than one functional category. 
   & $r_{\text{multi}}$ \\
\midrule
Edge type
 & Adjacent edge ratio   & $e_{\text{adj}}  = \card{\{(u,v) : \text{type} = \text{`adjacent'}\}} / \card{E}$ & $e_{\text{adj}}$ \\
 & Door edge ratio       & $e_{\text{door}} = \card{\{(u,v) : \text{type} = \text{`door'}\}} / \card{E}$     & $e_{\text{door}}$ \\
 & Corridor edge ratio   & $e_{\text{corr}} = \card{\{(u,v) : \text{type} = \text{`corridor'}\}} / \card{E}$ & $e_{\text{corr}}$ \\
\end{longtable}
}
 
\noindent{\footnotesize\textit{Notes.} Features are standardized before model fitting. Entropy is computed using the natural logarithm with a small constant ($10^{-10}$) added to avoid $\log(0)$. For LoG~1 graphs, each aggregated node is assigned the dominant function label of its constituent fine-level elements when computing function class ratio and entropy. All three edge type ratios are retained to preserve the full distributional information across edge types in a form directly usable by both linear and nonlinear models.}
 
{\footnotesize
\setlength{\tabcolsep}{4pt}
\renewcommand{\arraystretch}{1.3}
\begin{longtable}{@{}>{\raggedright\arraybackslash}p{2.2cm} >{\raggedright\arraybackslash}p{2.8cm} >{\raggedright\arraybackslash}p{8.0cm} >{\raggedright\arraybackslash}p{1.8cm}@{}}
\caption{Node-level feature set for multi-granularity layout graphs.}\label{tab:node-features}\\
\toprule
\textbf{Feature Group} & \textbf{Feature Name} & \textbf{Definition / Formula} & \textbf{Notation} \\
\midrule
\endfirsthead
\toprule
\textbf{Feature Group} & \textbf{Feature Name} & \textbf{Definition / Formula} & \textbf{Notation} \\
\midrule
\endhead
\bottomrule
\multicolumn{4}{r}{\textit{(continued on next page)}}\\
\endfoot
\bottomrule
\endlastfoot
Topology
 & Clustering coefficient
   & $C_v = 2 t_v / (k_v (k_v - 1))$, where $t_v$ = number of triangles through $v$; $C_v = 0$ if $k_v < 2$
   & $C_v$ \\
\midrule
Degree
 & Raw degree            & $k_v$ = number of edges incident to node $v$ & $k_v$ \\
 & Normalized degree     & $\hat{k}_v = k_v / \max\{k_i : i \in V\}$      & $\hat{k}_v$ \\
 & \textbf{K-core number}
   & $\mathrm{norm\_core}_v = \mathrm{core}(v) / \max\{\mathrm{core}(i) : i \in V\}$, where $\mathrm{core}(v)$ is the maximum $k$ such that $v$ belongs to the $k$-core subgraph
   & $\mathrm{norm\_core}_v$ \\
\midrule
Centrality
 & \textbf{Eccentricity} & $\varepsilon_v = \max_{u \in V} d(v, u) / D$, normalized by graph diameter $D$; $\varepsilon_v = 0$ for isolated nodes & $\varepsilon_v$ \\
 & Betweenness centrality& $C^b(v) = \sum_{s \neq v \neq t} \sigma(s, t \mid v) / \sigma(s, t)$; normalized to $[0,1]$ & $C^b(v)$ \\
 & Closeness centrality  & $C^c(v) = (\card{V}-1) / \sum_{u \neq v} d(u,v)$; normalized to $[0,1]$ & $C^c(v)$ \\
\midrule
Function
 & Known function one-hot
   & $\varphi_v$: one-hot encoding of $\text{func}(v)$ for non-masked nodes; zero vector for masked (target) nodes.
   & $\varphi_v$ \\
\midrule
Composition
 & Node size
   & $n_{\text{members},v} = \card{\{f_i : f_i \text{ merged into } v\}}$. Number of fine-level nodes aggregated into $v$; $n_{\text{members},v} = 1$ for fine-grained graphs.
   & $n_{\text{members},v}$ \\
\midrule
Edge type
 & Door edge count
   & $n^{\text{door}}_v = \card{\{u : (v,u) \in E, \text{type} = \text{`door'}\}}$. Number of door-type edges incident to $v$.
   & $n^{\text{door}}_v$ \\
 & Corridor edge count
   & $n^{\text{corr}}_v = \card{\{u : (v,u) \in E, \text{type} = \text{`corridor'}\}}$. Number of corridor-type edges incident to $v$. Note: $n^{\text{adj}}_v = k_v - n^{\text{door}}_v - n^{\text{corr}}_v$ is omitted as it is linearly redundant with degree.
   & $n^{\text{corr}}_v$ \\
\midrule
\textbf{Location}
 & \textbf{Spatial $x$}  & $\hat{x}_v = (x_v - \min_i x_i) / (\max_i x_i - \min_i x_i) \in [0,1]$. Normalized centroid $x$-coordinate within each floor plan. & $\hat{x}_v$ \\
 & \textbf{Spatial $y$}  & $\hat{y}_v = (y_v - \min_i y_i) / (\max_i y_i - \min_i y_i) \in [0,1]$. Normalized centroid $y$-coordinate within each floor plan. & $\hat{y}_v$ \\
\end{longtable}
}
 
\noindent{\footnotesize\textit{Notes.} Path features are not applicable for node-level. Features are standardized before input. \textbf{Bold} $=$ new node-level feature with no graph-level counterpart.}

Figure~\ref{fig:class-heatmap} reports class-wise performance for the zone function prediction task, showing that Categories 4 (Lounge) and 6 (Self-study) are essentially unpredictable in the five-class setting.

\begin{center}
\includegraphics[width=\linewidth]{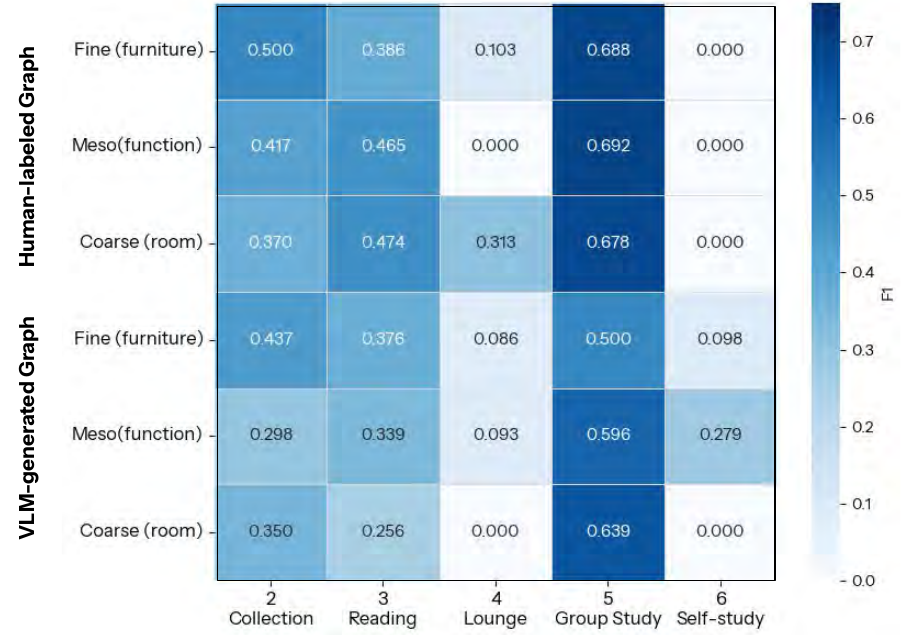}
\captionof{figure}{Class-wise F1 scores for zone function prediction (GINE).}
\label{fig:class-heatmap}
\end{center}

\begin{table}[!htbp]
\centering
\caption{Zone function prediction performance under the 5-class setting (Macro F1).}
\label{tab:appendix-results}
\footnotesize
\setlength{\tabcolsep}{6pt}
\renewcommand{\arraystretch}{1.15}
\begin{threeparttable}
\begin{tabular}{llccc}
\toprule
\textbf{Graph type} & \textbf{Model} & \textbf{LoG~3 (fine)} & \textbf{LoG~2 (meso)} & \textbf{LoG~1 (coarse)} \\
\midrule
\multirow{4}{*}{Human} & GINE      & 0.321 & 0.350            & / \\
                       & GCN       & 0.299          & 0.297            & / \\
                       & GraphSAGE & 0.313          & \textbf{0.369*}  & / \\
                       & GAT       & 0.272          & 0.251            & / \\
\cmidrule(lr){1-5}
\multirow{4}{*}{VLM}   & GINE      & 0.296          & 0.303            & / \\
                       & GCN       & 0.277          & 0.280            & / \\
                       & GraphSAGE & 0.290          & 0.326            & / \\
                       & GAT       & 0.260          & 0.238            & / \\
\bottomrule
\end{tabular}
\begin{tablenotes}[flushleft]\footnotesize
\item \textbf{Bold} $=$ best performance; \textbf{Bold $+$ *} $=$ recommended under the performance--complexity trade-off.
\end{tablenotes}
\end{threeparttable}
\end{table}
\end{document}